\pdfoutput=1

\documentclass[11pt]{article}

\usepackage[]{acl}

\usepackage{times}
\usepackage{latexsym}

\usepackage[T1]{fontenc}
\usepackage[utf8]{inputenc}

\usepackage{microtype}

\usepackage{inconsolata}

\usepackage{graphicx}

\usepackage{hyperref}       
\usepackage{url}            
\usepackage{booktabs}       
\usepackage{amsfonts}       
\usepackage{nicefrac}       
\usepackage{microtype}      
\usepackage{xcolor}         
\usepackage{microtype}
\usepackage{inconsolata}
\usepackage{amsmath,amsfonts,bm}
\usepackage{graphicx}
\usepackage{multirow}
\usepackage{makecell}
\usepackage{array}
\usepackage[para]{threeparttable}
\usepackage{booktabs}
\usepackage{tabularx}
\usepackage{xspace}
\usepackage{caption}
\usepackage{subcaption}
\usepackage{longtable}
\usepackage{hyperref}
\usepackage{bbm}
\usepackage{bm}
\usepackage{enumitem}
\usepackage{algorithm}
\usepackage{algpseudocode}
\usepackage{geometry}
\usepackage{comment}
\usepackage[colorinlistoftodos]{todonotes}
\usepackage{tablefootnote}
\usepackage{color}
\usepackage{tcolorbox}
\usepackage{tikz}
\usepackage{tabularray}
\usepackage[normalem]{ulem}

\usepackage[T1]{fontenc}
\usepackage[utf8]{inputenc}
\usepackage{listings}
\usepackage{xspace}
\definecolor{ForestGreen}{RGB}{34,139,34}

\title{Understanding and Enhancing Mamba-Transformer Hybrids \\ for Memory Recall and Language Modeling}

\author{
  Hyunji Lee$^{\hspace{.1em}{\boldsymbol{\mathcal{U}}}}$\thanks{\hspace{1mm}Work was done during internship at Tencent AI Lab, Bellevue.} \quad
  Wenhao Yu$^{\hspace{.1em}\boldsymbol{\tau}}$ \quad
  Hongming Zhang$^{\hspace{.1em}\boldsymbol{\tau}}$ \quad
  Kaixin Ma$^{\hspace{.1em}\boldsymbol{\tau}}$ \quad \\
  \bf Jiyeon Kim$^{\hspace{.1em}\boldsymbol{\kappa}}$ \quad 
  Dong Yu$^{\hspace{.1em}\boldsymbol{\tau}}$ \quad
  Minjoon Seo$^{\hspace{.1em}\boldsymbol{\kappa}}$ \quad \\
    \\
  $^{\mathcal{U}\hspace{.1em}}$UNC Chapel Hill \quad 
  $^{\tau\hspace{.1em}}$Tencent AI Lab\quad
  $^{\kappa\hspace{.1em}}$KAIST AI \quad 
   \\
}

\begin{document}
\maketitle
\begin{abstract}
Hybrid models that combine state space models (SSMs) with attention mechanisms have shown strong performance by leveraging the efficiency of SSMs and the high recall ability of attention. However, the architectural design choices behind these hybrid models remain insufficiently understood. In this work, we analyze hybrid architectures through the lens of memory utilization and overall performance, and propose a complementary method to further enhance their effectiveness. We first examine the distinction between sequential and parallel integration of SSM and attention layers. Our analysis reveals several interesting findings, including that sequential hybrids perform better on shorter contexts, whereas parallel hybrids are more effective for longer contexts. We also introduce a data-centric approach of continually training on datasets augmented with paraphrases, which further enhances recall while preserving other capabilities. It generalizes well across different base models and outperforms architectural modifications aimed at enhancing recall. Our findings provide a deeper understanding of hybrid SSM-attention models and offer practical guidance for designing architectures tailored to various use cases.
Our findings provide a deeper understanding of hybrid SSM-attention models and offer practical guidance for designing architectures tailored to various use cases\footnote{Code in \href{https://github.com/amy-hyunji/mamba-transformer-hybrids}{mamba-transformer-hybrids}}.

\end{abstract}
\section{Introduction}
Recent advances in state-space models~(SSMs), such as Mamba~\citep{mamba}, have shown strong performance in language modeling, particularly in long-context tasks, while offering significantly greater efficiency than traditional Transformer~\citep{vaswani2017attention} architectures~\citep{mamba2, waleffe2024empirical, zuo2024falcon}.
However, unlike Transformers, which maintain a dynamically growing key-value (KV) cache to attend to all previous tokens, SSMs compress past information into a fixed-size hidden state, limiting their ability to model long-term dependencies and recall distant context~\citep{park2024mambaformer, glorioso2024zamba}.
To address this, recent work has explored \textit{hybrid architectures}~\citep{hymba2024, ren2024samba, park2024mambaformer} that integrate attention with SSMs, aiming to leverage the strengths of both: combining the expressive, high capacity memory of attention with the efficiency of SSM computation.

Despite promising results, there remains a limited understanding of how different architectural design choices affect performance in these hybrid models, and what specific roles SSM and attention components play.
In this work, we aim to fill this gap by systematically analyzing the following three research questions:
(\emph{RQ1}) \textbf{Aggregation Strategies}: How do different ways of combining SSMs and attention affect performance and efficiency?
(\emph{RQ2}) \textbf{Component Roles}: What are the respective contributions and characteristics of SSMs and attention layers in hybrid models?
(\emph{RQ3}) \textbf{Data-Centric Enhancements}: Can performance be further improved through data-centric methods, beyond architectural design alone?

\begin{figure*}[t!]
    \begin{minipage}[b]{1.0\textwidth}
    \centering
    \includegraphics[width=\linewidth]{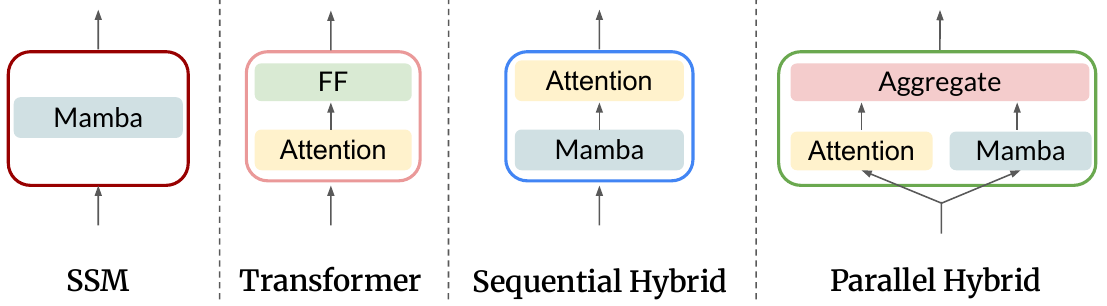}
    \end{minipage}
\caption{
Comparison of different architectural designs: SSM, Transformer, Sequential Hybrid, and Parallel Hybrid. Each architecture consists of stacked \textit{blocks} that incorporate Mamba and Attention layers. The key difference lies in how these layers are arranged: \textcolor{brown}{SSM} uses only Mamba layers, \textcolor{pink}{Transformer} uses only Attention layers, while the hybrid models combine both. \textcolor{blue}{Sequential Hybrid} stacks Mamba and Attention layers within each block, whereas \textcolor{ForestGreen}{Parallel Hybrid} applies them in parallel and aggregates their outputs. Feedforward~(FF) layers are omitted in the hybrid models for clarity, as it varies by design.
}
\label{fig:arch_comparison}
\end{figure*}

To investigate the first two questions~(\emph{RQ1}, \emph{RQ2}), we conduct extensive pretraining experiments on 17 models spanning pure SSMs, Transformer, and hybrid variants~(Figure~\ref{fig:arch_comparison}). 
Prior work often uses inconsistent training and evaluation setups, making fair comparison difficult. 
We therefore design a unified experimental setup that standardizes training and evaluation, enabling a controlled analysis of individual components and architectural choices.
All models share the same configurations, differing only in their core block design (SSM or attention).
We evaluate them across three axes: language modeling, commonsense reasoning, and memory recall.
Our analysis shows a strong correlation between long-context language modeling and commonsense reasoning, but weaker links to memory recall.
These results suggest that focusing solely on language modeling or reasoning benchmarks, as in prior work~\citep{glorioso2024zamba, lieber2024jamba, ren2024samba}, may miss critical aspects of memory performance. 
\textit{Our study fills this gap by providing a comprehensive and standardized evaluation.}

using our unified evaluation, we analyze \textit{how aggregation strategies (sequential or parallel) affect performance and the roles of ssm and attention components}. 
sequential hybrids, where one component processes input before the other, excel on short-context tasks because aligned representation spaces promote stable training. 
however, this alignment can limit expressiveness. 
in contrast, parallel hybrids keep separate embedding spaces and fuse outputs later, enabling greater representational diversity and stronger long-context performance. 
Among them, the parallel variant with a merge-attention layer, which attends over the outputs of the Mamba and the attention layers to produce a fused representation, achieves the strongest overall results.

Beyond architecture, we explore a \textit{data-centric approach} to improve memory recall (\emph{RQ3}). 
While previous works often finetune models on synthetic tasks like Needle-in-a-Haystack~(NIAH)~\citep{niah}, which boosts recall but often harms performance on other metrics. 
To mitigate this, we show that continued training with paraphrased sentences, drawn from a distribution similar to the pretraining data, enhances recall with minimum or no degradation in commonsense reasoning.
Compared to other datasets such as UltraChat~\citep{ding2023enhancing}, Based~\citep{arora2024just}, or NIAH, this strategy achieves the best trade-off. 
Notably, it outperforms architectural methods aimed at enhancing recall, such as DeciMamba~\citep{benkish2024decimambaexploringlengthextrapolation} (+12.7 avg), and generalizes well across a range of base models, scaling up to 2.8B parameter model.

\section{Preliminary} \label{sec: preliminary}
In this section, we share details of how the Mamba layer from recent SSM models and the Attention layer in the Transformer differ, an overview of prior works on hybrid models, and outline our experimental architectures.

\paragraph{Mamba and Attention layers} \label{subsec: mamba_attention}
Both Mamba and attention layers transform an input sequence into an output sequence using a transformation matrix, but differ in how they process inputs.
Mamba layers update a recurrent hidden state sequentially, incorporating one token at a time as a compressed summary of past inputs. In contrast, attention layers process the entire sequence simultaneously, attending to all preceding tokens to model dependencies.
These approaches involve trade-offs: Mamba offers linear-time computation but may struggle with long-range dependencies, while attention layers capture such dependencies more effectively at the cost of quadratic time and memory.
See Appendix~\ref{app_subsec: mamba_attention} for details and equations.

\paragraph{Hybrid Models} \label{subsec: hybrid_models}
To leverage the strengths of SSMs and attention, recent works have proposed hybrid architecture that integrate both components~\citep{hymba2024, ren2024samba, park2024mambaformer}. 
These models outperforms non-hybrid models, especially in long-context language modeling compared to attention-only models and recall performance compared to pure SSMs.

Recent hybrid models vary along four design axes: (1) SSM layer type: Mamba is the most common~\citep{mamba, ren2024samba, hymba2024, glorioso2024zamba}, though alternatives like DeltaNet have also been effective~\citep{yang2025gated}.
(2) Layer ratio: A 1:1 SSM-to-attention ratio is typical~\citep{hymba2024, ren2024samba}, though some prefer more SSM layers for efficiency~\citep{glorioso2024zamba, lieber2024jamba}.
(3) Attention type: To retain efficiency, many use SWA\footnote{SWA restricts attention to a fixed-size window around each token, improving scalability over full attention.}\citep{ren2024samba, yang2025gated}, combine SWA with full attention\citep{hymba2024}, or use full attention alone~\citep{glorioso2024zamba}.
(4) Integration strategy: Sequential fusion is most common~\citep{park2024mambaformer, ren2024samba, yang2025gated}, but parallel fusion is also explored~\citep{hymba2024}.

In this work, as our focus is on understanding affect of how to combine SSMs with attention layers and analyzing the role of each components in hyrid model performance, we focus on the fourth axes and keep other design choices fixed based on the recent strong baselines~\citep{ren2024samba, hymba2024, yang2025gated}: (1) using Mamba as the SSM component, (2) a 1:1 ratio of attention to SSM layers, and (3) using SWA~\citep{Beltagy2020Longformer} as attention layer. See Appendix~\ref{app: related_works} for more related works.

\paragraph{Architectural Designs of Hybrid Blocks} \label{subsec: hybrid_block}

To analyze various hybrid model configurations, we design a set of \textit{hybrid models}, each combining Mamba and Attention layers. These blocks are stacked to build the full model~(Figure~\ref{fig:arch_comparison}).
Our designs vary along two main axes: (1) the integration strategy and (2) the placement of feed-forward (FF) layers.
For integration, we explore: \textbf{sequential hybrid} where one layer’s output feeds into the other, with two variants (Mamba $\rightarrow$ SWA and SWA $\rightarrow$ Mamba) and \textbf{parallel fusion} where both layers receive the same input, and their outputs are aggregated using one of several methods (simple averaging~\citep{hymba2024}, a trainable projection layer~\citep{behrouz2024titans}, or a trainable merge-attention layer).
Also, given that FF layers play an important role in Transformer models~\citep{geva2020transformer, Meng2022LocatingAE}, we also experiment with the effect of different FF placements.

\section{Designing a Unified Experimental Setup}
While various works have proposed and demonstrated the effectiveness of hybrid models, their results are often difficult to compare to each other due to differences in training procedures, evaluation metrics, and the absence of released checkpoints.
To enable fair and comprehensive analysis, in this section, we introduce a unified experimental setup to re-evaluate multiple models within this consistent framework of training (Section~\ref{subsec: training_setup}) and evaluation (Section~\ref{subsec: evaluation_setup}).
We observe that some prior works often overlook key metrics, which can obscure a model's overall performance, underscoring the need for extensive evaluation over multiple axes to understand model performance.

\begin{figure}[t!]
    \begin{minipage}[b]{0.48\textwidth}
    \centering
    \includegraphics[width=\linewidth]{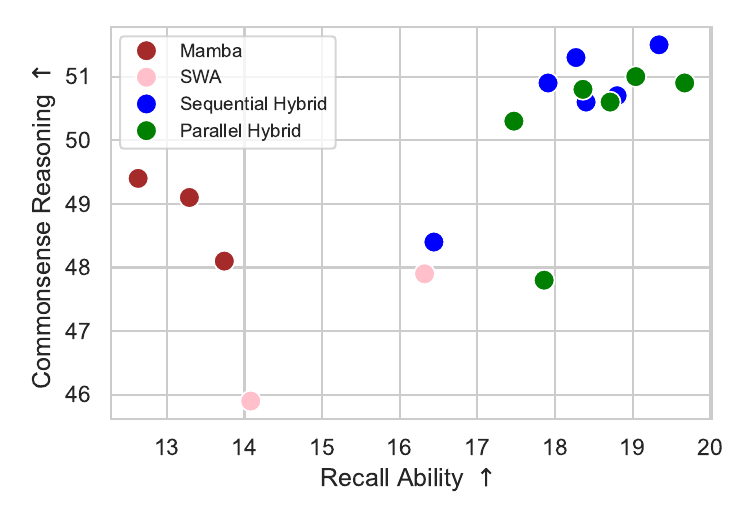}
    \end{minipage}
\caption{
    Comparison of different model architectures on Commonsense Reasoning (y-axis) vs. Recall Ability (x-axis).
    Commonsense Reasoning and Recall Ability are measured using answer accuracy.
    The models compared included \textcolor{brown}{Mamba-only}, \textcolor{pink}{SWA-only}, \textcolor{blue}{Hybrid (Sequential)}, and \textcolor{ForestGreen}{Hybrid (Parallel)}. 
    For details of each model, see Figure~\ref{app_fig:scatter_names} in Appendix~\ref{app_subsec: correlation}.
}
\label{fig:correlation}
\vspace{-1em}
\end{figure}

\begin{figure*}[ht!]
    \centering
    \begin{minipage}[t]{0.48\linewidth}
        \centering
        \includegraphics[width=\linewidth]{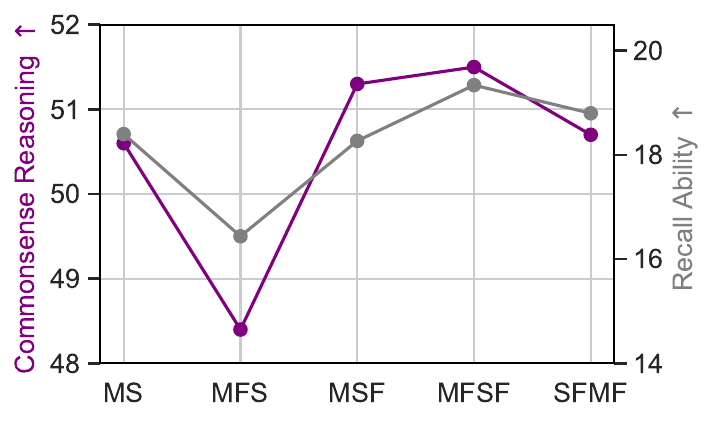}
        \subcaption{}
        \label{fig:sequential}
    \end{minipage}
    \hfill
    \begin{minipage}[t]{0.48\linewidth}
        \centering
        \includegraphics[width=\linewidth]{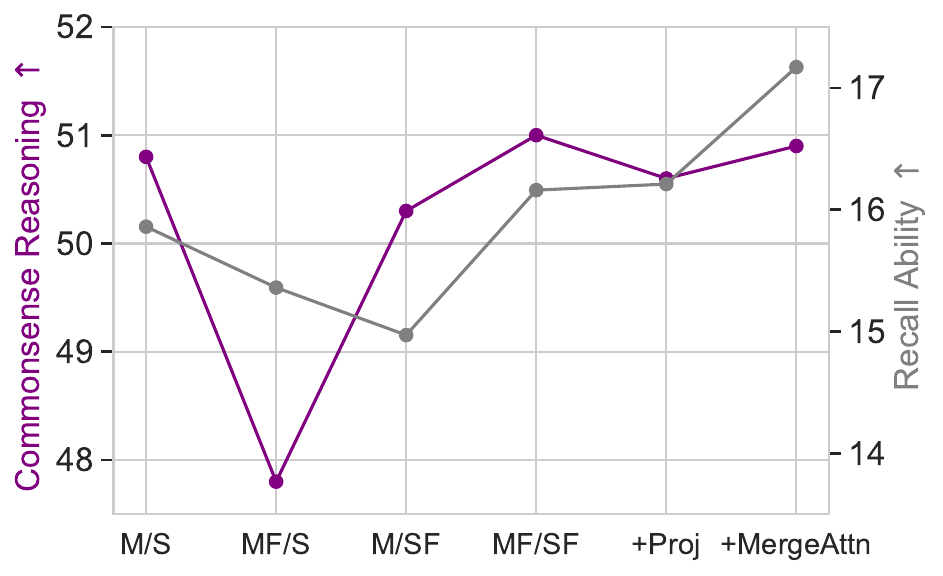}
        \subcaption{}
        \label{fig:parallel}
    \end{minipage}
    \caption{
    Performance comparison of \textcolor{purple}{commonsense reasoning accuracy} and \textcolor{gray}{recall ability} across different model architectures. (\textbf{a}) Results for sequential models. (\textbf{b}) Results for parallel models. For further details, refer to the first paragraph of Section~\ref{sub: architecture_result}.
    }
    \label{fig:sequential_parallel}
    \vspace{-1em}
\end{figure*}

\subsection{Training} \label{subsec: training_setup}
We follow widely adopted training setups from recent works, primarily based on \citet{ren2024samba}, which provides detailed implementation code.
All models are trained from scratch on 100B tokens from the SlimPajama dataset~\citep{cerebras2023slimpajama}. 
Model sizes are kept consistent across architectural variants: approximately 430M parameters for base models and 1.3B for larger ones.
All models use the same hyperparameters: batch size of 512, sequence length of 4K, learning rate of 4e-4, weight decay of 0.1, window size of 2k for SWA, and the AdamW optimizer~\citep{loshchilov2017decoupled}.

\subsection{Evaluation} \label{subsec: evaluation_setup}
\paragraph{Setup}
We evaluate hybrid models across three axes: (1) long-context language modeling, (2) commonsense reasoning, and (3) memory recall, following previous works on hybrid models.
For language modeling, we report perplexity on the SlimPajama validation set using 16k-token sequences.
Commonsense reasoning is assessed by averaging accuracy across five standard benchmarks: LAMBADA-OpenAI~\citep{radford2019language}, HellaSwag~\citep{zellers2019hellaswag}, PIQA~\citep{Bisk2020}, ARC-Easy~\citep{allenai:arc}, and Winogrande~\citep{ai2:winogrande}. 
Recall ability is evaluated over average of eight datasets in Based benchmark~\citep{arora2024just}, using the evaluation protocol of \citet{yang2025gated}. We further group them into short- and long-context subsets to study the influence of context length on recall performance. Details are in Appendix~\ref{app_subsec: dataset}.

\paragraph{Correlation Between Evaluation Axes}

We investigate how the three evaluation axes, language modeling, commonsense reasoning, and memory recall, relate across different architectural choices.
We find that \textbf{strong performance on reasoning or language modeling does not necessarily imply strong memory recall.} 
While there is some positive correlation, it is relatively weak.
Specifically, the pearson correlation coefficient between language modeling and commonsense reasoning is high (0.814), whereas recall correlates modestly with reasoning (0.697) and even less with language modeling (0.542).
These trends are also visualized in Figure~\ref{fig:correlation}, which shows the correlation between recall ability (x-axis) and reasoning (y-axis).
Notably, the clustering of models with similar architectures (indicated by color) suggests that architectural design has a greater impact on recall performance than overall reasoning ability.
These findings highlight that prior works, which evaluate models solely on language modeling or reasoning benchmarks~\citep{glorioso2024zamba, lieber2024jamba, ren2024samba}, need a more comprehensive evaluation including memory-intensive tasks to more accurately assess model capabilities.

\section{How Does the Architectural Design Affect Model Performance?}
In this section, we present our experimental results across various model architectures (Section~\ref{sub: architecture_result}) and provide a detailed analysis of their structural design (Section~\ref{sub: architectural_analysis}).

\subsection{Results} \label{sub: architecture_result} 
Figure~\ref{fig:sequential_parallel} compares commonsense reasoning and recall performance across various \textit{block} designs in both sequential and parallel model architectures. 
In both subfigures, the x-axis represents different block configurations.
\textsc{M} indicates a Mamba layer, \textsc{S} a Sliding Window Attention (SWA) layer, and \textsc{F} a feed-forward (FF) layer. For example, \textsc{MFSF} represents a block with Mamba, FF, SWA, and FF layers in that order.
In parallel models~(Figure~\ref{fig:parallel}), `|' denotes parallel branches (e.g., M|SF means Mamba on one side and SWA+FF on the other).  
Aggregation strategies are defined as follows: \textsc{+Proj} uses a trainable projection layer; \textsc{+MergeAttn} uses a trainable attention module, similar to the cross-attention layer in encoder-decoder models, but using Mamba's output embeddings as the Key and Value; the remaining variants use simple mean averaging. 
See Appendix~\ref{app_subsec: 430M} for detailed performance and Appendix~\ref{app_subsec: hybrid_nonhybrid} for comparison between hybrid and non-hybrid models. 

\paragraph{Impact of SWA and Mamba Layer Order on Sequential Hybrid Performance}

We investigate how the order of SWA and Mamba layers affect sequential hybrid performance by comparing two configurations: MFSF (Mamba before SWA) and SFMF (SWA before Mamba). As shown in Figure~\ref{fig:sequential}, MFSF consistently outperforms SFMF across tasks.
This suggests that placing the Mamba layer first helps the model capture global dependencies early, while placing SWA first may bottleneck performance due to its limited attention window.
However, when analyzing recall performance by context length (Figure~\ref{app_fig:sequential.short_long} in Appendix~\ref{app_subsec: seq_short_long}), SFMF performs better on shorter contexts. We attribute this to SWA effectively approximating full attention when the input length is within its window, enabling strong local representations that Mamba can refine.
In summary, SFMF may benefit short-context tasks, but MFSF, architecture used in \citet{ren2024samba}, offers superior overall performance. We therefore adopt MFSF as our default sequential model architecture.

\paragraph{Effect of Aggregation Method in Parallel Hybrids Performance}

We study how different aggregation layers for combining SWA and Mamba output embeddings affect hybrid model performance. 
As shown in Figure~\ref{fig:parallel}, we evaluate three strategies: \textsc{+Both}, \textsc{Proj}, and \textsc{MergeAttn}.
\textsc{MergeAttn} achieves the best overall performance, particularly in long-context language modeling (see Appendix~\ref{app_subsec: parallel_agg} for more results). We thus use \textsc{MergeAttn} as the representative parallel model in subsequent analysis.
Simple averaging (\textsc{+Both}) performs well on commonsense reasoning, consistent with observation in \citet{hymba2024}, but we observe that it underperforms on recall; for strong recall, especially with long contexts, trainable aggregation methods like \textsc{Proj} and \textsc{MergeAttn} are more effective.

\paragraph{Sequential models excel in short contexts, parallel models excel in long ones}

\begin{figure}[t!]
    \centering
    \begin{minipage}[t]{0.48\textwidth}
    \includegraphics[width=\linewidth]{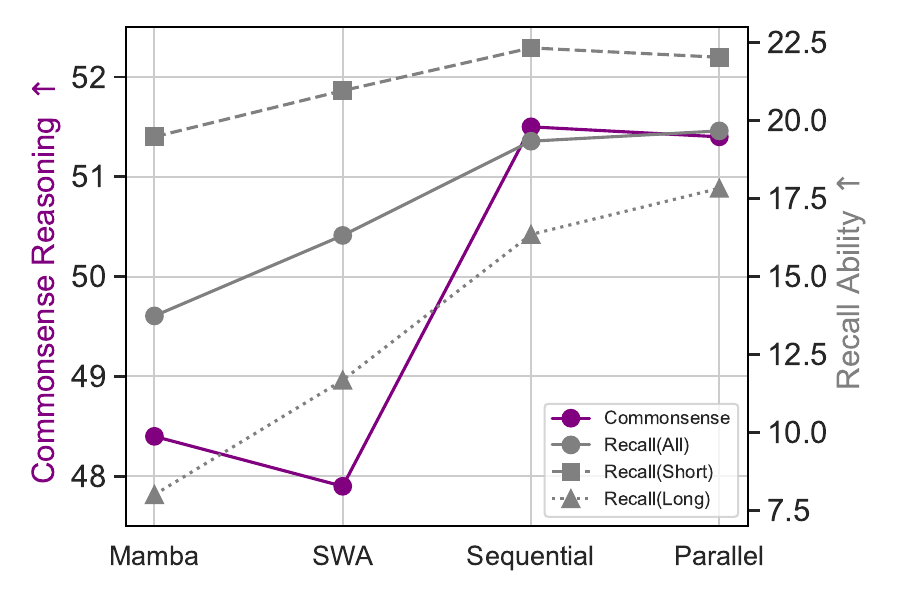}
    \caption{Performance of best performing models from each architecture in \textcolor{purple}{commonsense reasoning} and \textcolor{gray}{recall ability}, where divided by length of context.} 
    \label{fig:short_long_comparison}
    \end{minipage}
    \vspace{-1em}
\end{figure}

When comparing recall performance of sequential and parallel models, we observe that sequential models tend to perform better in relatively shorter contexts whereas parallel combinations general show superior performance in longer contexts (Figure~\ref{fig:short_long_comparison}).
We hypothesize that this trend arises from the differing degrees of interaction between the SWA and Mamba components. 
As parallel models has less interaction between Mamba and SWA components, it prevents from collapsing into a shared mode of producing overly similar hidden states. It instead encourages each component to retain its distinct representational strength. In Section~\ref{subsec: parallel_seq}, we provide empirical evidence supporting this hypothesis.

\paragraph{Impact of Adding Feed-Forward Layers on Hybrid Model Performance}

Feed-formward~(FF) layers play an important role in transformers~\citep{geva2020transformer, Meng2022LocatingAE}, but their effect on hybrid models remains less explored.
We find that adding FF layers to only one component, either Mamba or SWA, degrades performance in both sequential and parallel settings, while improvements appear only when FF layers are added to \textit{both} components.
We hypothesize that this degradation arises from feature misalignment: it is especially harmful in parallel architectures, where components maintain distinct representations and make aggregation harder, whereas sequential models integrate features into a shared space, mitigating some of these issues.
This drop is particularly high when adding FFNs to Mamba, likely because its final layer ($C$ in Equation~\ref{eq: lrnn}) already functions similarly to an MLP~\citep{sharma2024locating}, making additional FFNs redundant or even detrimental. This aligns with prior findings that FFNs benefit SWA but not Mamba~\citep{mamba}.

\paragraph{Generalization to 1.3B}
Trends observed at the 430M scale generally hold at 1.3B. 
Hybrid models consistently outperform non-hybrids. 
Among sequential hybrids, MFSF outperforms MS. 
In parallel setups, merge-attention as an aggregation layer shows higher performance, especially for long-context recall.
Overall, merge-attention mechanisms show strong performance.
Sequential hybrids excel in short-context settings, while parallel hybrids perform better with longer contexts. See Appendix~\ref{app_subsec: 1.3B} for detailed results.

\subsection{Analysis}  \label{sub: architectural_analysis}
\begin{figure}[t!]
    \centering
    \begin{minipage}[t]{0.48\textwidth}
    \includegraphics[width=\linewidth]{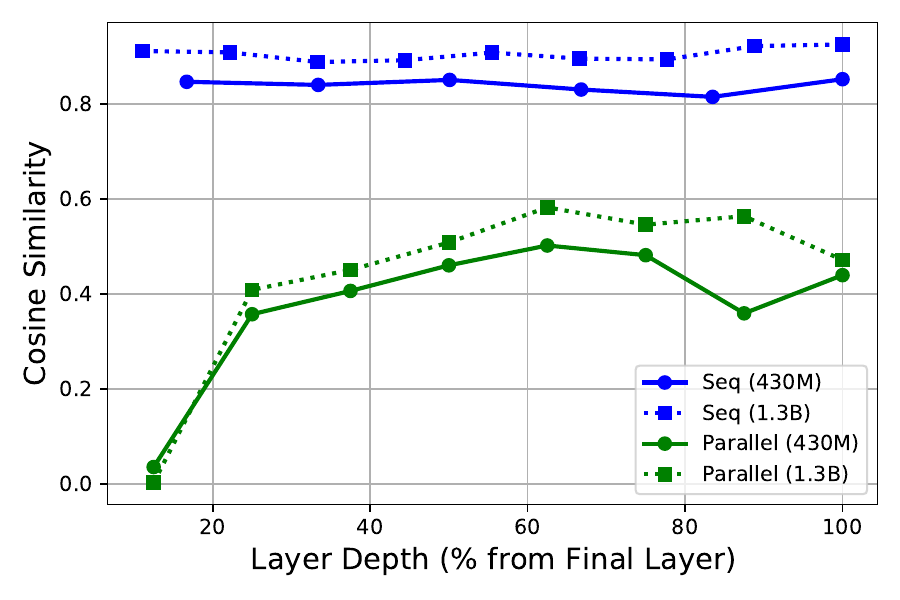}
    \caption{Cosine similarity between output embeddings of aligned SWA and Mamba layers (y-axis), plotted against layer depth, measured as percentage distance from the final layer (x-axis).} 
    \label{fig:cosine_similarity}
    \end{minipage}
    \vspace{-1em}
\end{figure}
\paragraph{Similarity between SWA and Mamba Output Embeddings in Hybrid Models} \label{subsec: parallel_seq}

To better understand the interaction between SWA and Mamba in hybrid models, we analyze the cosine similarity of their output embeddings across block depths, aligned by position from the final block (Figure~\ref{fig:cosine_similarity}).
Sequential hybrids show high similarity, especially in the larger 1.3B model, because outputs from one component feed into the next, naturally aligning their representations. 
Parallel hybrids show much lower similarity, particularly in early and middle layers, as both components process inputs independently and fuse outputs later. 
We hypothesize that this structural difference shapes performance: sequential hybrids benefit from stable, aligned representations for commonsense reasoning and short-context tasks but struggle with long-context reasoning. 
In contrast, parallel hybrids produce more diverse representations and, though sensitive to aggregation strategy, can outperform on complex long-context tasks when effectively trained.
More analysis in Appendix~\ref{app_subsec: similarity btw output}.

\begin{figure*}[ht!]
    \begin{minipage}[b]{1.0\textwidth}
    \centering
    \includegraphics[width=0.82\textwidth]{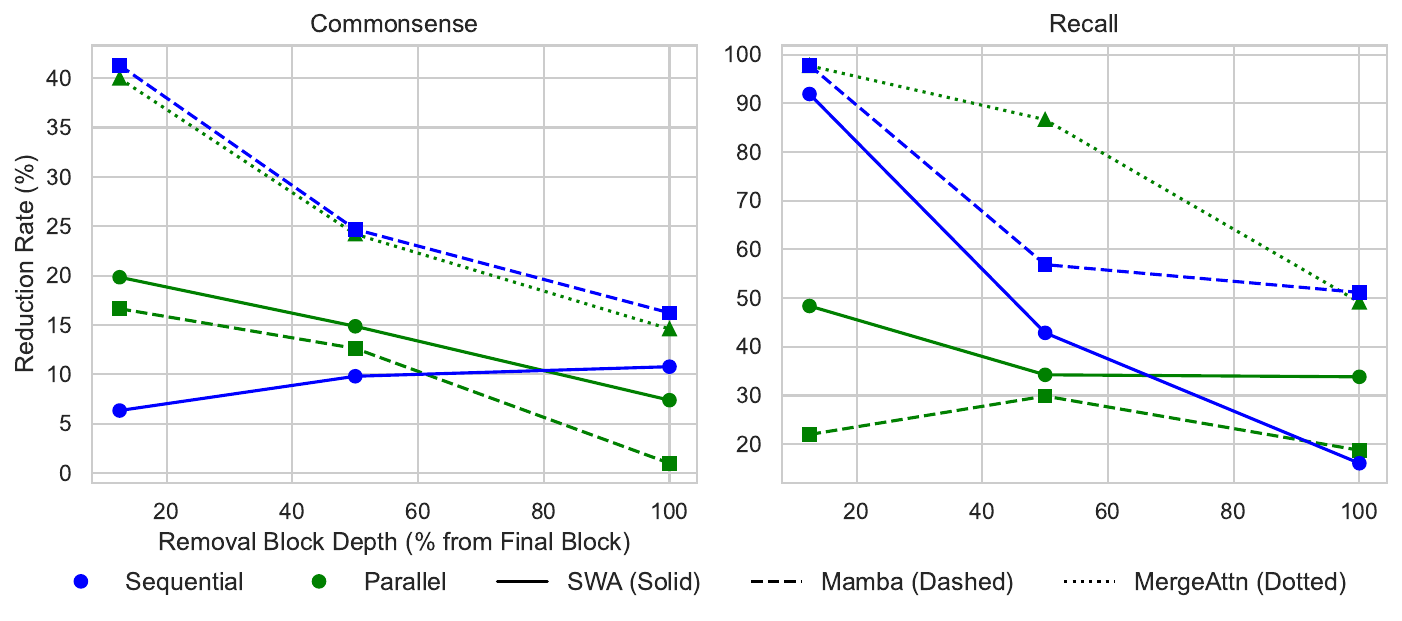}
    \end{minipage}
\caption{
Performance degradation (y-axis) on commonsense (left) and recall (right) tasks as a function of the removed block's relative position from the final block (x-axis).
} 
\label{fig:removal_layer}
\vspace{-1em}
\end{figure*}

\paragraph{Identifying Critical Components in Hybrid Blocks} 

Figure~\ref{fig:removal_layer} shows performance degradation on commonsense (left) and recall (right) tasks when removing blocks by depth.
Removing the first block causes the steepest drop, up to 90\% on recall tasks, highlighting the crucial role of early layers. 
We further examine the importance of subcomponents within each block.
In sequential models, the first subcomponent is most critical because it shapes the feature space, and later components align to it.
In parallel models, the aggregation layer is most critical as it must merge the distinct representation spaces from Mamba and SWA, while either path alone can still infer the input distribution. 
See Appendix~\ref{app_subsec: critical_layer} for further discussion.

\begin{figure}[t!]
    \centering
    \begin{minipage}[t]{0.48\textwidth}
   \includegraphics[width=\linewidth]{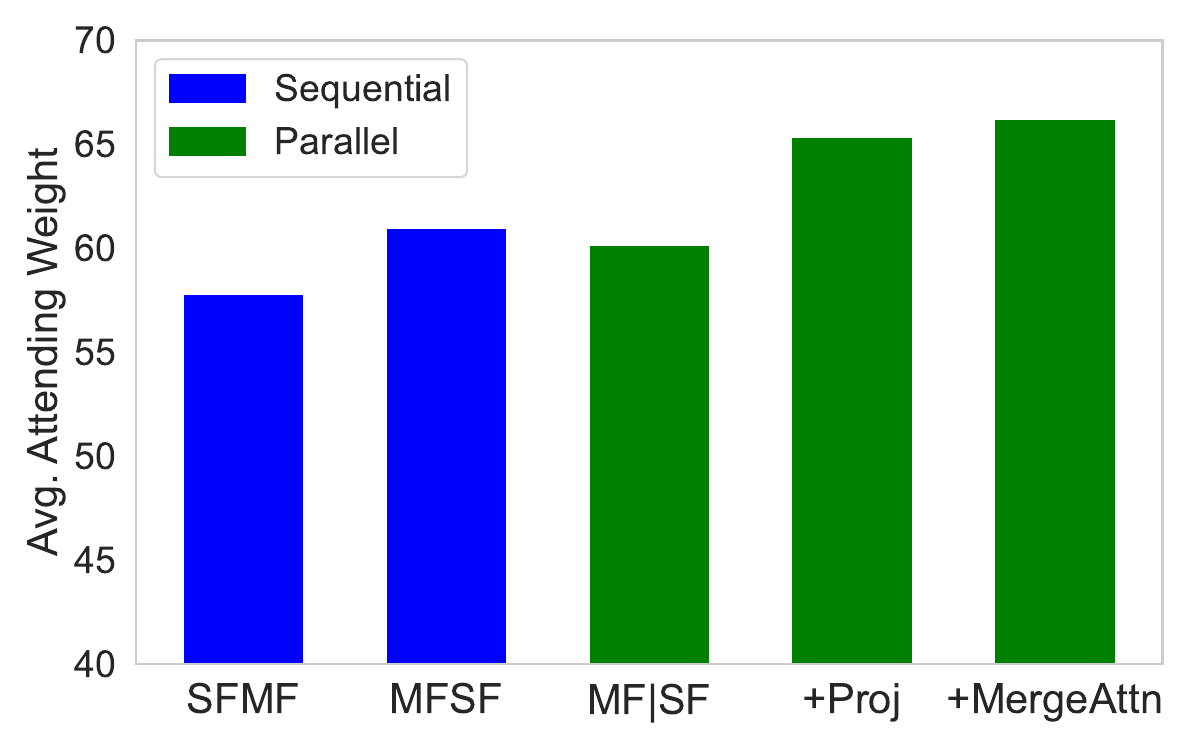}
    \caption{Average attending weight across different model architectures. Higher values indicate that the model attends more strongly to previous information.} 
    \label{fig:avg_attention}

    \end{minipage}
    \vspace{-1em}
\end{figure}

\paragraph{Understanding the performance gains of \textsc{MergeAttn}}
Among the various configurations, parallel hybrids using an attention layer that merges output embeddings of SSM and attention achieve the best performance.
To understand why these models tend to perform strongly, we analyze the models on how much each token is influenced by prior tokens, following the method in \citet{benkish2024decimambaexploringlengthextrapolation}; higher value indicates that they exhibit stronger attention to previous tokens. 
As shown in Figure~\ref{fig:avg_attention}, models with merge-attention show the highest average attention weights, suggesting that their improved performance arises because the Mamba layers effectively capture global dependencies, which the merge-attention mechanism then leverages to integrate information. See Appendix~\ref{app_subsec: attn_weight} for more details of the calculation.

\section{Dataset Strategy to Enhance Recall}

We show that continually training models on datasets with paraphrased contexts, drawn from a distribution similar to the pretraining dataset, improves recall without sacrificing commonsense reasoning.
Previous work focused on improving recall through \textit{architectural changes}, such as hybrid models. 
Here, we investigate a \textit{data-centric approach}, aiming to complement and further enhance these architectural advances.

Section~\ref{subsec5: exp_setup} describes how we construct the training dataset and train the model.
Section~\ref{subsec5: exp_result} shows that models trained on our dataset achieve the best trade-off between recall and reasoning, outperforming other dataset choices and DeciMamba (which introduces architectural changes) across scales up to 2.8B parameters.
Section~\ref{subsec5: analysis} analyzes design factors such as input length, dataset size, and model choice, demonstrating through extensive experiments that our simple approach generalizes well and consistently improves performance.

\subsection{Experimental Setup} \label{subsec5: exp_setup}
\paragraph{Paraphrasing Method}
We construct a paraphrased dataset using a subset of the training corpus~(SlimPajama), based on the hypothesis that the data should remain close in distribution to the original pretraining corpus to prevent degrading existing performance. 
To control the density of paraphrased content, we divide the data into 1k-token chunks. For each chunk, we use LLaMA 3.1–8B\footnote{We use the released model from Hugging Face: \texttt{meta-llama/Meta-Llama-3-8B-Instruct}} to generate factual question-answer (QA) pairs. Following \citet{arora2024just}, we convert these QA pairs into cloze-style paraphrased sentences.
This yields pairs of the form (1k-token chunk, paraphrased sentence). 
To construct a training dataset, we concatenate multiple chunks and insert the corresponding paraphrased sentence at a random position following the chunk it was derived from. 
Based on the constructed dataset, we run a filtering process based on three criteria: (1) the model fails to generate a valid question and answer pair, (2) the generated answer is not present in the corresponding paragraph, or (3) the model fails to convert the example into a cloze-style task. 
See Appendix~\ref{app_subsec: paraphrase_filtering} for more details.

\paragraph{Training Details}
After the initial pretraining phase,\footnote{We also experimented with incorporating the paraphrase dataset during pretraining. However, we observed a degradation in performance when doing so (see Appendix~\ref{app_subsec: paraphrase_order}).} we continue training the model using several different datasets, including recall-intensive datasets such as NIAH and SQuAD from Based, widely used SFT dataset UltraChat~\citep{ding2023enhancing}, and our paraphrased dataset.
Following the setup of \citet{benkish2024decimambaexploringlengthextrapolation}, we train the models using a batch size of 32, a learning rate of 1e-4 for 10 epochs. We conduct experiments with both hybrid models and Mamba-only models.

\subsection{Results} \label{subsec5: exp_result}

\begin{figure}[t!]
    \begin{minipage}[b]{1.0\linewidth}
    \centering
    \includegraphics[width=1.0\linewidth]{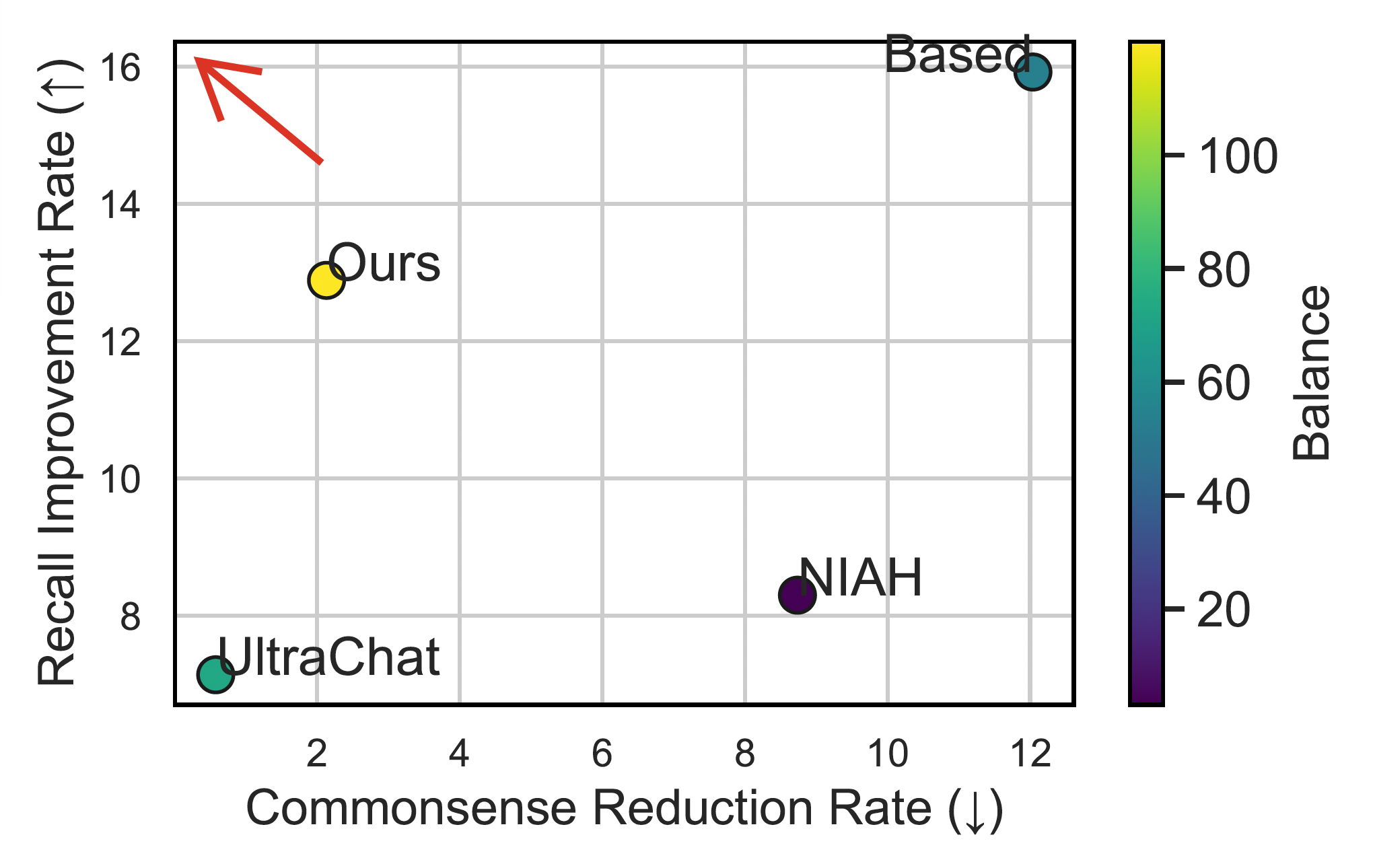}
    \end{minipage}
\caption{
The upper-left region (indicated by the \textcolor{red}{red} arrow) represents the optimal balance between recall improvement and commonsense degradation. 
}
\label{fig:seq.hybrid_tradeoff}
\vspace{-1em}
\end{figure}

\paragraph{Our Dataset Strikes the Best Balance}
Figure~\ref{fig:seq.hybrid_tradeoff} shows the balance between commonsense degradation and recall gains\footnote{In this section, we exclude SQuAD from Based when computing average recall, as it is part of the training data.} for the 430M sequential hybrid model (MFSF) when trained on various datasets including NIAH~\citep{niah}, UltraChat~\cite{waleffe2024empirical}, or SQuAD dataset from Based benchmark~\citep{arora2024just}. 
Models trained on our paraphrased SlimPajama dataset consistently achieve the best balance.
We attribute this to: (1) its alignment with the original pretraining distribution, preserving baseline performance; and (2) paraphrased content promoting the model to retain and utilize previous context.
In contrast, recall-focused datasets like NIAH and Based significantly harm commonsense performance, while UltraChat offers only modest recall improvements.
Appendix~\ref{app_subsec: balance_hybrid} provides detailed performance. Similar patterns hold for Mamba models, including the released 2.8B version (Appendix~\ref{app_subsec: balance_mamba}).

\begin{figure}[t!]
    \centering
    \begin{minipage}[t]{0.48\textwidth}
   \includegraphics[width=\linewidth]{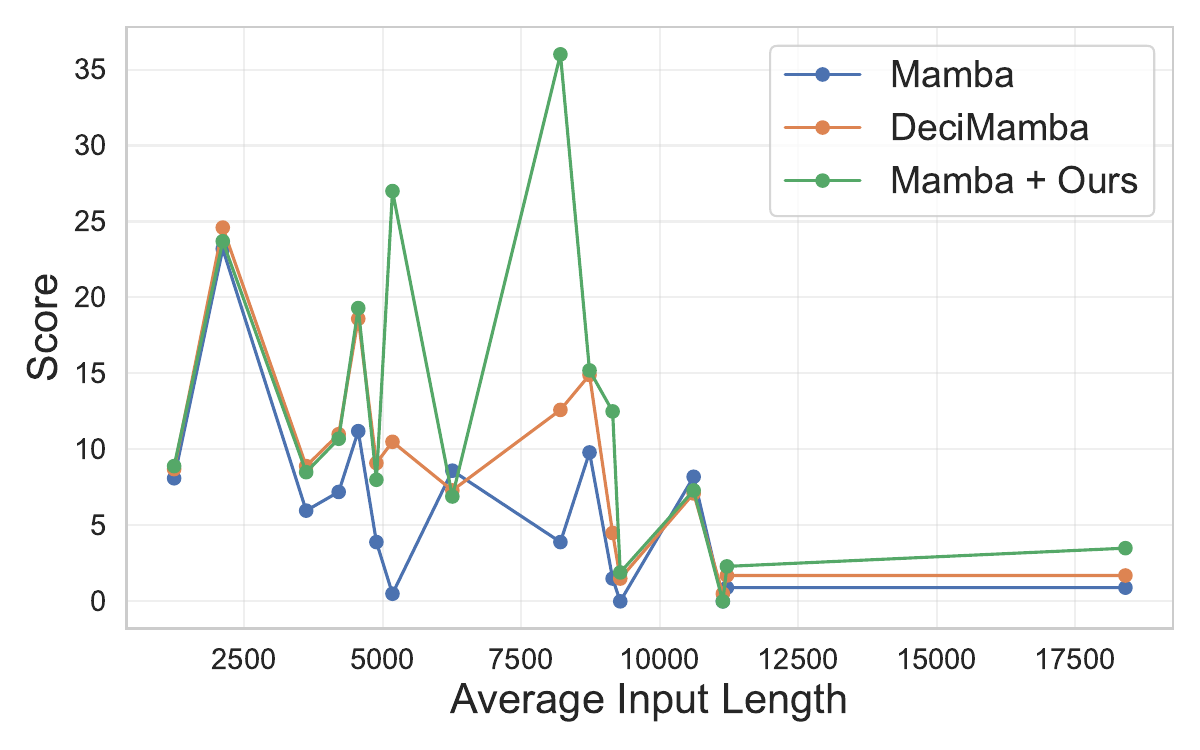}
    \caption{Performance (y-axis) of Mamba, DeciMamba, and Mamba trained with our dataset across LongBench datasets with varying input lengths (x-axis).} 
    \label{fig:decimamba}
    \end{minipage}
    \vspace{-1em}
\end{figure}

\paragraph{Comparison with DeciMamba}

We investigate whether a \textit{data-centric} approach can outperform architectural modifications by comparing our approach with DeciMamba~\citep{benkish2024decimambaexploringlengthextrapolation}, which enhances recall by discarding less important tokens. 
Across 16 datasets in LongBench~\citep{bai2023longbench} using Mamba-2.8B, our approach achieves an average of +12.7 points overall, with particularly strong gains in QA tasks (+8.1 on average). 
As shown in Figure~\ref{fig:decimamba}, our approach tends to consistently outperform DeciMamba on medimum and long input lengths.
These results suggest that our data-centric approach is not only complementary to architectural change but can also show strong standalone performance.
Full results are provided in Appendix~\ref{app_subsec: decimamba}.

\subsection{Analysis}  \label{subsec5: analysis}

To understand the benefit and affect of such approach, we analyzed over various design choices.

\paragraph{Generalizes to Various Base Models}
We observe that \textbf{our method generalizes across different released variants of Mamba-2.8B}. Continual training yields performance gains for the base model (+1.7 in commonsense, +6.5 in recall), as well as for instruction-tuned (+1.3 in commonsense, +3.6 in recall) and preference-aligned models (+3.4 in commonsense, +0.8 in recall). Improvements are generally more pronounced for the base model. See Appendix~\ref{app_subsec: basemodel} for model details and results.

\paragraph{Longer chunk sizes yield stronger results}

We observe that models trained on longer sequences tend to achieve lower reduction rates on commonsense tasks and substantially higher gains on recall tasks, especially on long-context tasks~(Figure~\ref{fig:dataloader_length} in Appendix)\footnote{All experiments used a fixed token count of 10M, discarding the final chunk if it does not align with the sequence length.}.
Training on shorter chunk sizes (e.g., 2k) tends to enhance performance on short-context recall but leads to high degradation on long-context tasks. 
This trend is robust across architectures, appearing in both sequential and parallel hybrids, and holds for different model sizes.
For detailed results, refer to Appendix~\ref{app_subsec: paraphrase_training_dataset_length}.

\paragraph{Performance improves as the size of the training dataset increases} \label{subsec: training_dataset_cnt}

We observe that training with larger datasets leads to clear gains in both commonsense reasoning and recall tasks: models trained with more tokens achieve a lower reduction rate on commonsense benchmarks and steadily higher improvements on recall performance\footnote{All experiments use a fixed chunk size of 4k}.
Performance grows with the amount of training data and begins to converge around 80M-100M tokens.
This trend holds across different model sizes and hybrid architectures.
More details are in Appendix~\ref{app_subsec: training_dataset_cnt}.

\section{Conclusion}

In this paper, we focus on two main aspects: (1) studying how different architectural design choices (sequential, parallel) affect performance in hybrid models and the roles of individual components (SSM and Attention layers), and (2) exploring data-centric approaches to further improve model’s recall ability.
Our findings show that sequential models offer stable training but are limited in expressiveness, while parallel architectures better preserve the unique characteristics of each component, often leading to stronger performance. In particular, parallel hybrid models with merge-attention-based aggregation consistently perform well.
We also demonstrate that continually pretraining the model on a paraphrased dataset effectively improves recall while maintaining overall model performance.

\section*{Limitations}
Due to computational constraints, we conducted our experiments on relatively small model scales, 430M and 1.3B parameters, trained with 100B tokens. Pretraining a 430M-parameter model on 8 A100 GPUs takes about one week, while a 1.3B-parameter model requires roughly two weeks, making it challenging to analyze larger-scale models. Notably, prior work on hybrid models has also primarily operated at similar scales~\citep{ren2024samba, hymba2024, yang2025gated}.
These resource limitations also restricted our ability to explore a broader range of hybrid model configurations and focus on the experimental setup described in the “Hybrid Model” section (Section~\ref{sec: preliminary}).
We leave more extensive analyses, such as incorporating additional components like gated DeltaNet, to future work.

\bibliography{custom}

\appendix
\newpage
\section{Preliminary}

\subsection{Mamba and Attention layers} \label{app_subsec: mamba_attention}
Given an input sequence $X$, both Mamba layers and attention layers transform it into an output sequence $Y$ via a transformation matrix $M$: $M_{\text{Mamba}}$ (Equation~\ref{eq: lrnn}) and $M_{\text{Attn}}$ (Equation~\ref{eq: attn}).
The key difference lies in how they process inputs.
Mamba layers update a recurrent hidden state $h_t$ sequentially, incorporating one token $x_t$ at a time. This hidden state serves as a compressed memory summarizing all past inputs.
In contrast, Attention layers process the entire input sequence at once, attending to all tokens up to the current position, thereby capturing dependencies without recurrence.
These design choices yield different trade-offs. Mamba is more computationally efficient due to its linear-time recurrence but may struggle with long-range dependencies. Attention layers, while effective at modeling token-wise relationships, incur quadratic time and memory complexity with sequence length.

\begin{flushleft}
\begin{equation}
Y_{\text{Mamba}} = M_{\text{Mamba}} X \quad \text{where}
\label{eq: lrnn}
\end{equation}
\begin{equation*}
Y_{\text{Mamba}, t} = C h_t,\quad h_t = A h_{t-1} + B x_t,\quad x_t \in X
\end{equation*}
\begin{equation}
Y_{\text{Attn}} = M_{\text{Attn}} X \quad \text{where}
\label{eq: attn}
\end{equation}
\begin{equation*}
Y_{\text{Attn}} = \text{softmax}\left( 
\frac{(W_Q X)(W_K X)^T}{\sqrt{d_k}} 
\right)(W_V X)
\end{equation*}
\end{flushleft}

\begin{figure*}[t!]
    \begin{minipage}[b]{1.0\textwidth}
    \centering
    \includegraphics[width=0.9\textwidth]{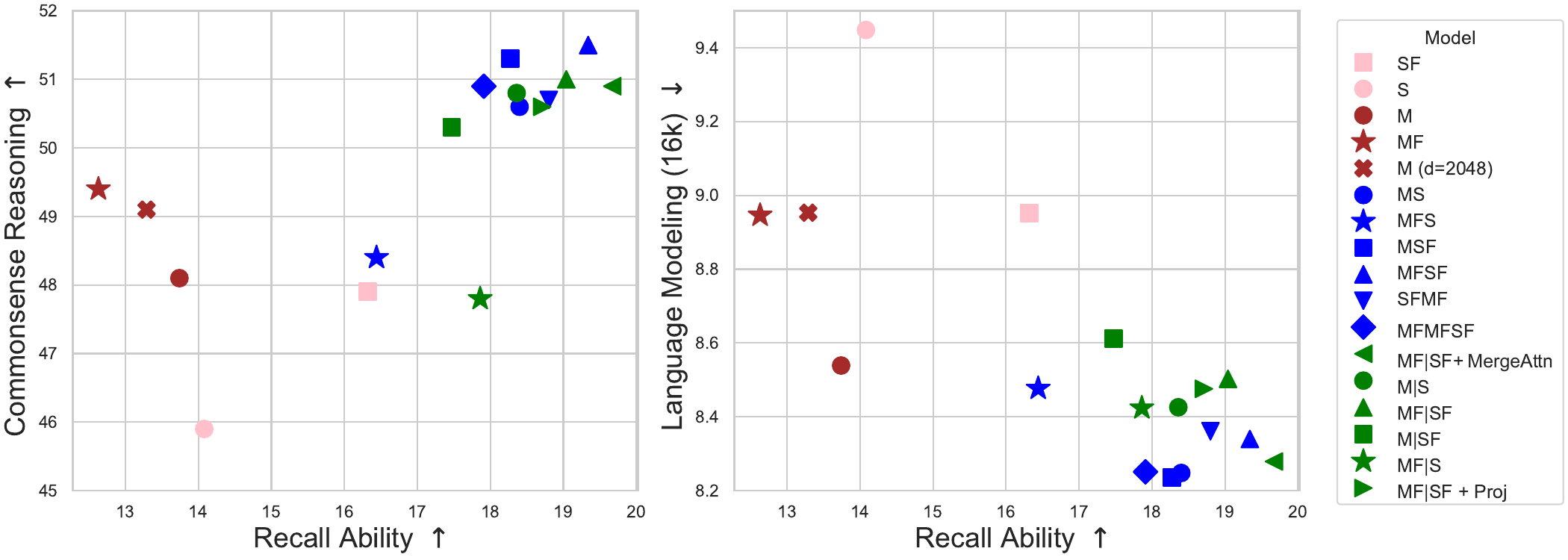}
    \end{minipage}
\caption{Comparison of detailed model architectures on Commonsense Reasoning (y-axis on left) and Language Modeling(y-axis on right) vs. Recall Ability (x-axis).
Commonsense Reasoning and Recall Ability are measured using answer accuracy.
The models compared included \textcolor{brown}{Mamba-only}, \textcolor{pink}{SWA-only}, \textcolor{blue}{Hybrid (Sequential)}, and \textcolor{ForestGreen}{Hybrid (Parallel)}.} 

\label{app_fig:scatter_names}
\end{figure*}

\section{Related Works} \label{app: related_works}

\subsection{Studies on SSMs}
Prior work has explored state space models (SSMs)~\citep{waleffe2024empirical, sharma2024locating}, primarily focusing on their performance in language modeling tasks, particularly their ability to handle long-context dependencies. However, these studies typically examine pure SSM architectures and do not consider hybrid models.
In this work, we conduct an empirical investigation of \textit{hybrid architectures} that combine SSM and attention layers. Our goal is to understand the source of their performance gains and the distinct roles played by each component.

\subsection{Recall Ability of Language Models}

Enhancing a model's recall ability, also referred to as grounding ability, is a critical aspect of language modeling, especially in scenarios where the model must answer questions based on a given context, maintain strong coherence across parts of a conversation or document, or perform consistent reasoning over extended texts~\citep{arora2024just, lee2023well}. This ability allows the model to retrieve relevant information accurately from given context, sustain contextual coherence, and generate factually grounded responses.

In this paper, we define recall ability as distinct from the general capability to model long contexts. Unlike next-token prediction, recall-intensive tasks require the model to retrieve specific values or answers from earlier in the context, demanding precise and accurate memory. Furthermore, evaluating recall ability is not limited to long-context tasks; it applies to any setting where exact retrieval from prior context is necessary.

Several studies have shown that SSM-based models often struggle with such recall-intensive tasks, as they must encode prior context into fixed-size hidden states. This architectural constraint leads to a bottleneck that limits their recall performance~\citep{park2024mambaformer, ren2024samba, hymba2024}.

\section{Designing a Unified Experimental Setup}

\subsection{Correlation Between Evaluation Axes} \label{app_subsec: correlation}
Figure~\ref{app_fig:scatter_names} shows the detailed configuration of each point in Figure~\ref{fig:correlation}.

\subsection{Dataset} \label{app_subsec: dataset}

We experiment over datasets from \citet{arora2024just} (Based benchmark) to calculate recall ability. Based benchmark is comprised of eight datasets: NQ~\citep{kwiatkowski2019natural}, TriviaQA~\citep{joshi2017triviaqa}, DROP~\citep{dua2019drop}, FDA~\citep{arora2023language}, SWDE~\citep{lockard-etal-2019-openceres}, and SQuAD~\citep{rajpurkar2018know}. The NQ dataset is further subdivided by input length into NQ-512, NQ-1024, and NQ-2048.
To compare the recall ability across different sequence lengths, we categorize the eight datasets into two groups: relatively short sequences (NQ-512, DROP, TriviaQA, SQuAD) and relatively long sequences (NQ-1024, NQ-2048, FDA, SWDE).
Using the LLaMA-2 tokenizer~\citep{touvron2023llama}, which was also used during training, the average input length is around 1k tokens for the short-sequence group and around 2.5k tokens for the long-sequence group.
The Slimpajama dataset and evaluation datasets are released under Apache 2.0 license. We used the datasets for research purposes.

\section{How does the architectural design affect model performance?}

\subsection{Performance at the 430M scale} \label{app_subsec: 430M}
\begin{figure}[t!]
    \begin{minipage}[b]{1.0\linewidth}
    \centering
    \includegraphics[width=0.9\linewidth]{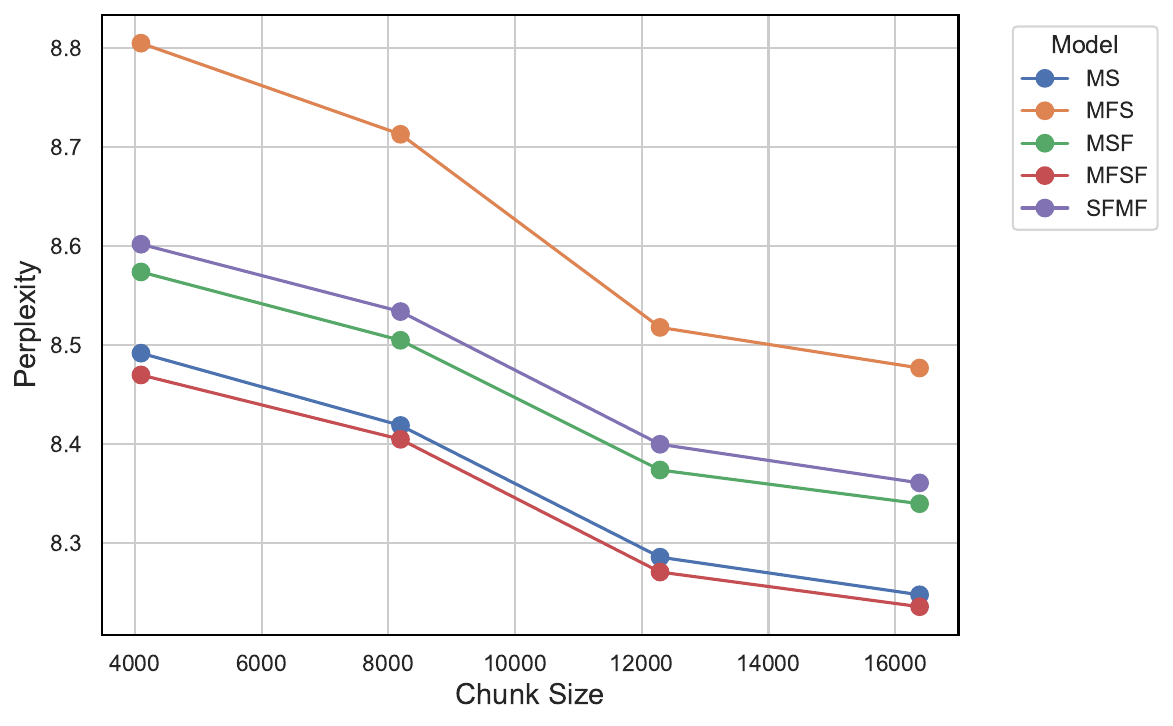}
    \end{minipage}
\caption{Peplexity over sequence length for sequential hybrids} 
\label{app_fig:seq.lm}
\end{figure}

Table~\ref{table: 430M} presents the performance of models at a 430M parameter scale.
Figures~\ref{app_fig:seq.lm} and~\ref{app_fig:parallel.lm} show the language modeling performance, measured in terms of perplexity on the SlimPajama validation set, for sequential and parallel hybrid models, respectively.

\begin{table*}[h]
\centering
\fontsize{6.5}{10} \selectfont
    \begin{tabular}{c|ccccc|c|cccccccc|c}
    \toprule
    &  \multicolumn{6}{c}{Commonsense Reasoning} |& \multicolumn{9}{c}{Recall Ability} \\ 
    \midrule
    Model Type& LAM. & Hella. & PIQA & ARC & Wino. & \textit{\textbf{Avg.}} & NQ-S & NQ-M & NQ-L & Drop & FDA & SWDE & TQA & SQD  &\textit{\textbf{Avg.}}\\
    \midrule
    M & 31.7 & 43.1 & 68.2 & 45.2 & 52.6 & 48.1  & 9.6	&8.7	&7.5	&11.4	&1.8&	14.1&	38.5&	18.4& 13.7\\
    MF & 34.2 & 41.6 & 68.6 & 51.8 & 51.0  & 49.4& 9.2&	8.8	&6.3&	10.4&	1.1	&12.3	&36.0&	17.0 & 12.6\\
    \midrule
    S & 30.6 & 37.6 & 64.6 & 45.3 & 51.3 & 45.9 & 9.9&9.2	&6.4	&12.4&	3.2&	18.4&	31.9&	13.4 & 13.1\\
    SF & 35.4 & 38.7 & 65.7 & 48.7 & 51.0 & 47.9&10.8	&7.9&	6.8	&12.3	&15.5&	16.5&	40.8&	20.0 & 12.6 \\
    \midrule
    \multicolumn{6}{l}{\textbf{Sequential Hybrid}} \\
    \midrule
    MS & 40.8 & 44.0 & 67.2 & 50.0 & 50.9 & 50.6 & 12.6	&12.2	&8.0	&11.6	&14.3	&26.4	&41.7	&20.6 &18.4\\
    MFS & 34.9 & 41.8 & 67.6 & 44.5 & 53.2 & 48.4 & 10.1	&8.8&	7.6&	11.8&	14.6	&21.6&	37.8&	19.3 & 16.4\\
    MSF & 39.0 & 43.3 & 67.9 & 52.5 & 53.8 & 51.3& 12.3&	11.5&	7.0&	11.7	&16.4&	24.8&	41.9&	20.6 & 18.3\\
    MFSF & 38.5 & 44.2 & 69.1 & 51.7 & 54.0 & \textbf{51.5}& 13.5	&12.3&	8.0&	11.2	&16.5	&28.6&	43.4&	21.2 & 19.3\\
    SFMF & 37.4 & 42.8 & 68.6 & 52.1 & 52.5 & 50.7&12.8	&11.6&	7.6&	12.2&	15.5&	25.6&	43.2&	22.0 & 18.8\\
    \midrule
    \multicolumn{6}{l}{\textbf{Parallel Hybrid}} \\
    \midrule
    M|S & 40.1 & 42.7 & 67.9 & 50.1 & 53.1 & 50.8 &11.5&	10.5	&7.3&	11.9&	16.2&	26.7&	42.8&	20.0 & 18.4\\
    MF|S & 24.5 & 42.5 & 68.3 & 52.3 & 51.7 & 47.8 &11.3&	11.3&	7.2&	11.6&	15.5&	25.6&	40.9&	19.6 & 17.9\\
    M|SF & 37.5 & 41.2 & 67.6 & 51.4 & 53.7 & 50.3 & 11.0	&10.0	&6.9&	10.9&	14.9&	24.5&	41.7&	19.9 & 17.5\\
    MF|SF (Avg) &  39.3 & 42.8 & 67.9 & 52.8 & 52.3 & 51.0 & 11.7	&11.9&	8.0	&12.3	&16.8&	28.0&	42.1&	19.9 & 18.8\\
    MF|SF (Proj) & 38.0 & 42.6 & 69.4 & 51.0 & 52.0 & 50.6 &11.9&12.4&	8.4&	12.7&	16.9&	28.6&	42.3&	20.7 & 19.2\\
    MF|SF (MergeAttn) & 39.3 & 44.3 & 69.0 & 51.9 & 52.3 & 51.4&12.8&	12.9	&9.0&	11.9&	17.7	&29.6&	43.1&	20.3 & \textbf{19.7}\\
    \bottomrule
    \end{tabular}
\caption
     {
     Model performance at the 430M scale. 
    Model Type: $M$ = Mamba, $S$ = SWA, $F$ = FF layer. The order reflects the design sequence within each block. In parallel hybrids, "|" denotes parallel branches (e.g., M|SF means Mamba on one side, SWA+FF on the other).
    Tasks: LAM. = LAMBADA-OpenAI, Hella. = HellaSwag, ARC = ARC-Easy, Wino. = Winogrande, NQ-S = NQ-512, NQ-M = NQ-1024, NQ-L = NQ-2048, TQA = TriviaQA, SQD = SQuAD.
    Bold indicates the highest average performance. In both cases, the best models use hybrid architectures with merge-attention.
     } 
\label{table: 430M}
\end{table*}

\subsection{Hybrid models outperform non-hybrid models in recall and commonsense reasoning} \label{app_subsec: hybrid_nonhybrid}
Results in Figure~\ref{fig:short_long_comparison} show the performance of non-hybrid models (Mamba, SWA) and hybrid models (Sequential, Parallel). 
We observe consistent gains in both commonsense reasoning and recall performance in hybrid models over non-hybrid ones in line with prior works~\citep{ren2024samba, hymba2024, park2024mambaformer}. 
Notably, we observe that recall shows a substantially larger improvement, with an average increase of 29.5\%, compared to a 7.3\% gain in commonsense reasoning. These improvements are especially prominent in long-context scenarios. In contrast, in short-context settings, performance differences are less pronounced, and hybrid models perform similarly to the SWA baseline.

\subsection{SWA as the Initial Component Improves Short Context Recall} \label{app_subsec: seq_short_long}
\begin{figure*}[t!]
    \begin{minipage}[b]{1.0\textwidth}
    \centering
    \includegraphics[width=0.9\textwidth]{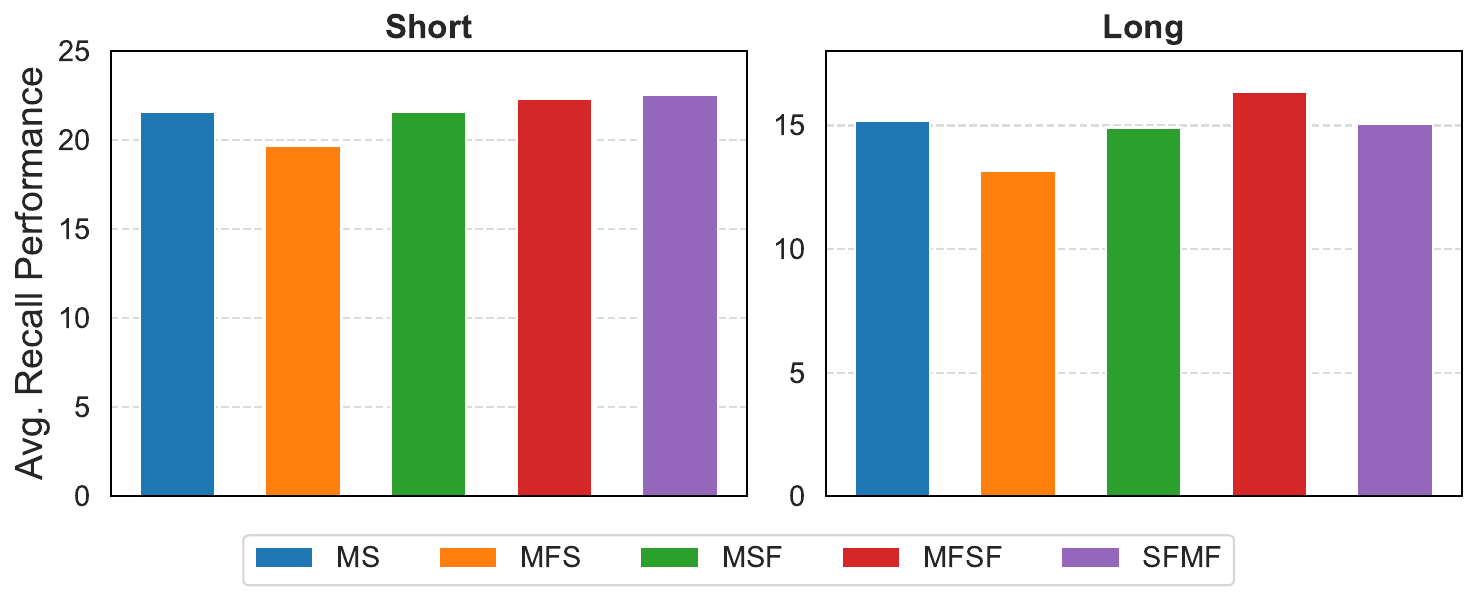}
    \end{minipage}
\caption{Comparison of average recall performance across short and long input contexts for sequential hybrids} 
\label{app_fig:sequential.short_long}
\end{figure*}

Figure~\ref{app_fig:sequential.short_long} presents the average recall performance for both short and long sequences in the sequential architecture. The SFMF configuration demonstrates stronger performance on shorter sequences. We hypothesize that this is because, in short contexts, the input length fits within the window size of the SWA module, allowing it to approximate full attention more effectively.

\begin{figure}[t!]
    \begin{minipage}[b]{1.0\linewidth}
    \centering
    \includegraphics[width=0.9\linewidth]{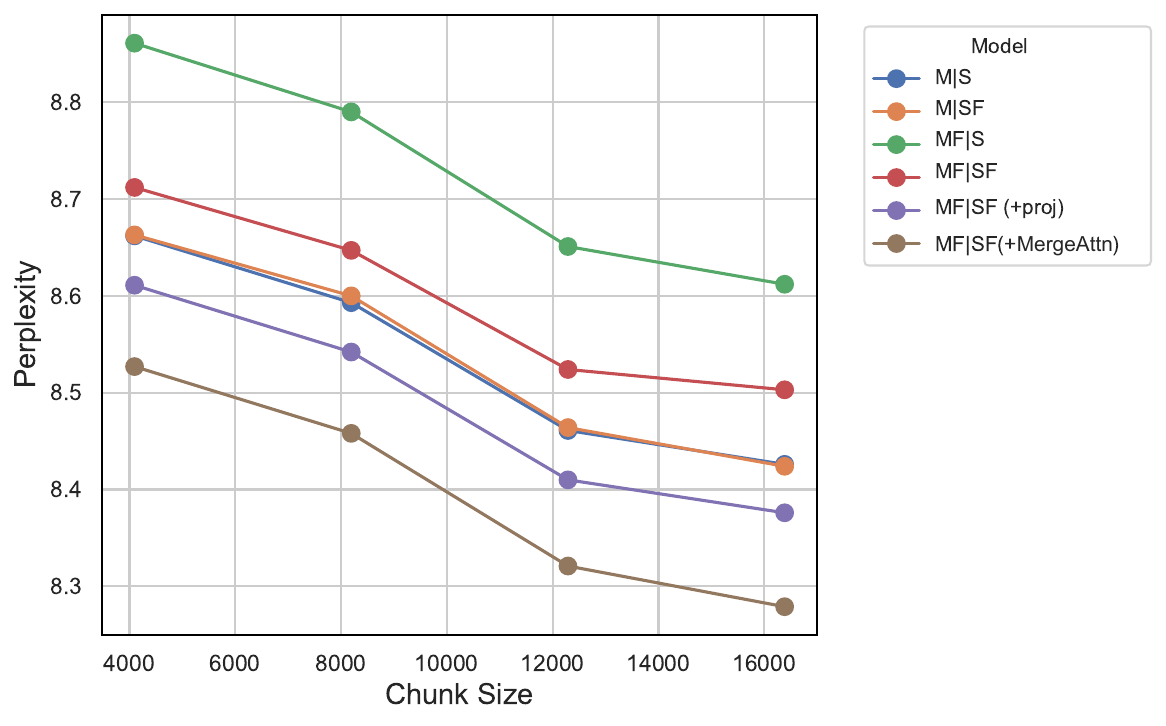}
    \end{minipage}
\caption{Peplexity over sequence length for parallel hybrids} 
\label{app_fig:parallel.lm}
\end{figure}

\begin{figure*}[t!]
    \begin{minipage}[b]{1.0\textwidth}
    \centering
    \includegraphics[width=0.9\textwidth]{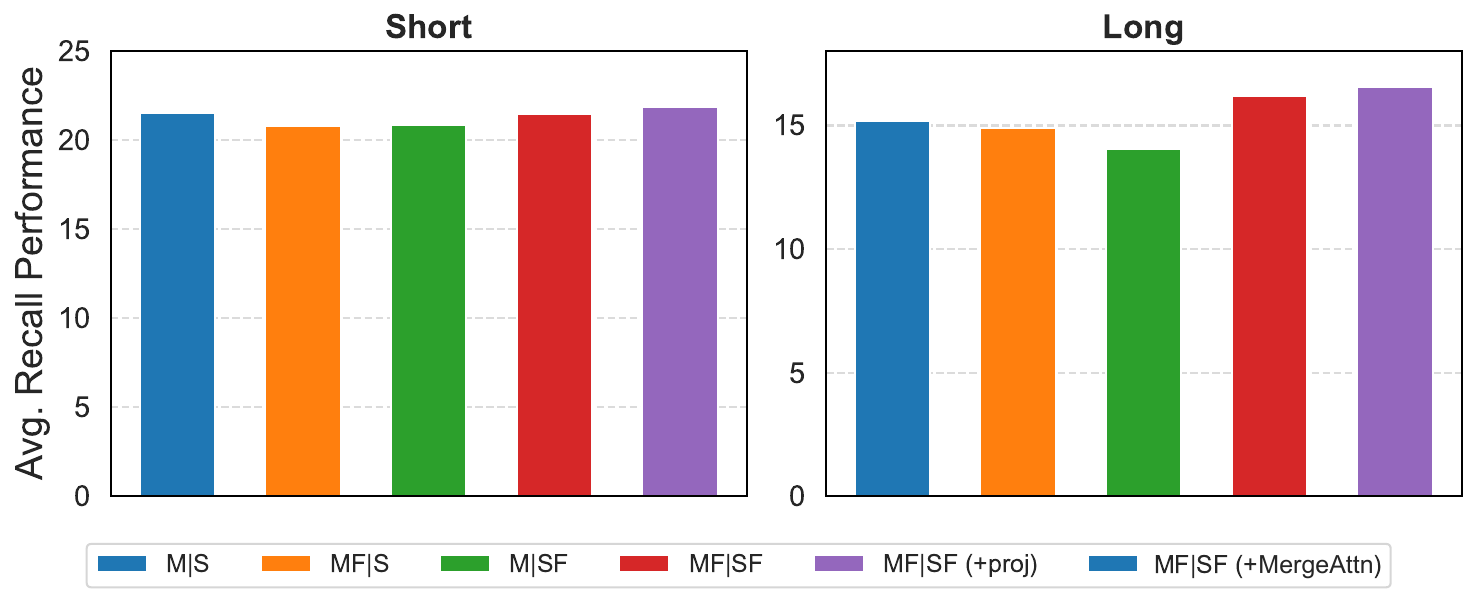}
    \end{minipage}
\caption{Comparison of average recall performance across short and long input contexts for parallel hybrids} 
\label{app_fig:parallel.short_long}
\end{figure*}

\subsection{Trainable Aggregation Layers Improve Performance on Long Contexts} \label{app_subsec: parallel_agg}

Figure~\ref{app_fig:parallel.lm} presents perplexity on the SlimPajama validation dataset across different chunk sizes. Models equipped with trainable aggregation layers, specifically MF|SF (+proj) and MF|SF (+MergeAttn), consistently outperform others across varying context lengths. These models show strong recall performance, with particularly notable improvements in longer ones (Figure~\ref{app_fig:parallel.short_long}).

\begin{figure*}[t!]
    \begin{minipage}[b]{1.0\textwidth}
    \centering
    \includegraphics[width=0.9\textwidth]{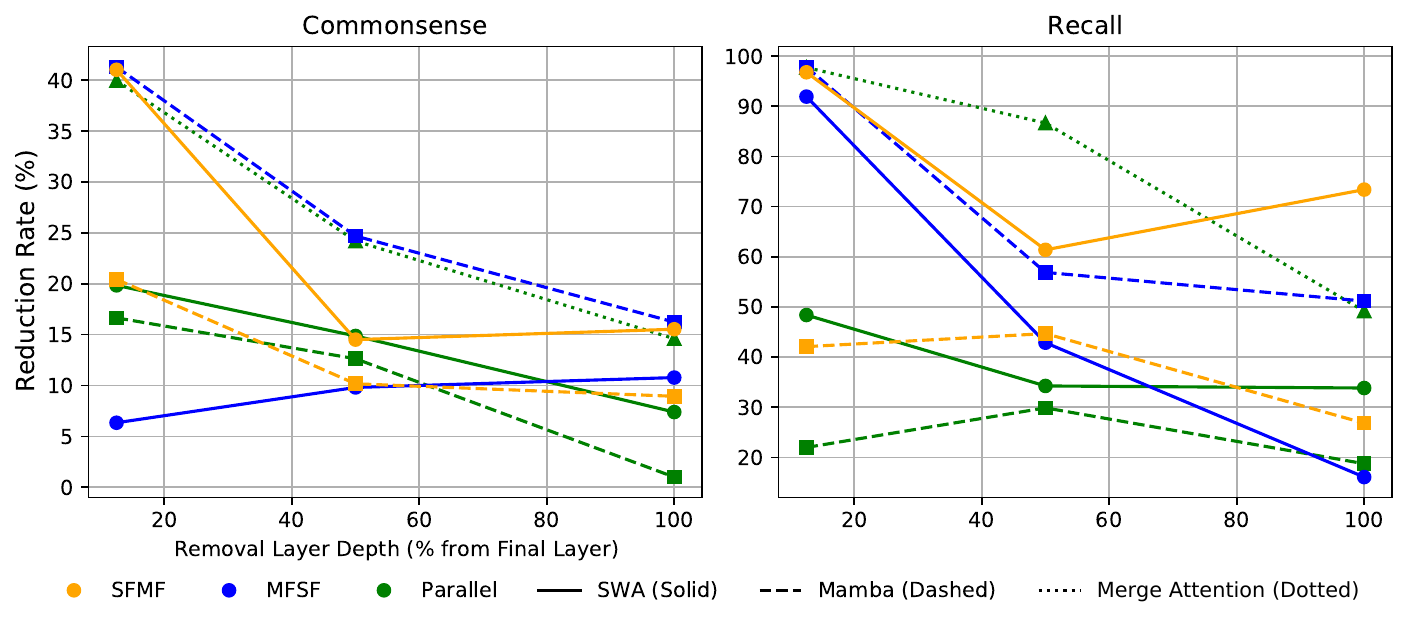}
    \end{minipage}
\caption{Performance degradation (y-axis) on commonsense (left) and recall (right) tasks as a function of the removed block's relative position from the final block (x-axis) for sequential(SFMF), sequential(MFSF) and parallel(+MergeAttn) architecture.
} 
\label{app_fig:removal_layer}
\end{figure*}

\subsection{Impact of Adding Feed-Forward Layers on Hybrid Model Performance}
Prior work has shown the importance of feed-forward~(FF) layers in transformers~\citep{geva2020transformer, Meng2022LocatingAE}. Thereby, we investigate their impact on hybrid models. Interestingly, adding FF layers to only one component, either Mamba or SWA, degrades performance in both sequential and parallel settings, and performance improves only when FF layers are added to both components.
In the sequential setup (Figure~\ref{fig:sequential}), the baseline \textsc{MS} outperforms \textsc{MFS} and \textsc{MSF}, but is lower than \textsc{MFSF}. Similarly, in the parallel setup (Figure~\ref{fig:parallel}), \textsc{M|S} show higher performance over \textsc{MF|S} and \textsc{SF|M} but lower than \textsc{SF|MF}.

We hypothesize that this degradation stems from feature misalignment. 
The effect is more pronounced in parallel architectures, where individual component characteristics are preserved, making it harder to aggregate misaligned features. 
In contrast, sequential models integrate features into a shared space, mitigating this effect. 
Also, the performance drop is especially large when adding FFNs to the Mamba layer, possibly because its final layer ($C$ in Equation~\ref{eq: lrnn}) already functions similarly to an MLP~\citep{sharma2024locating}, making an additional FFN redundant or even detrimental. This aligns with prior findings that FFNs benefit SWA but not Mamba~\citep{mamba}.

\begin{figure}[t!]
    \centering
    \begin{minipage}[t]{0.48\textwidth}
    \includegraphics[width=\linewidth]{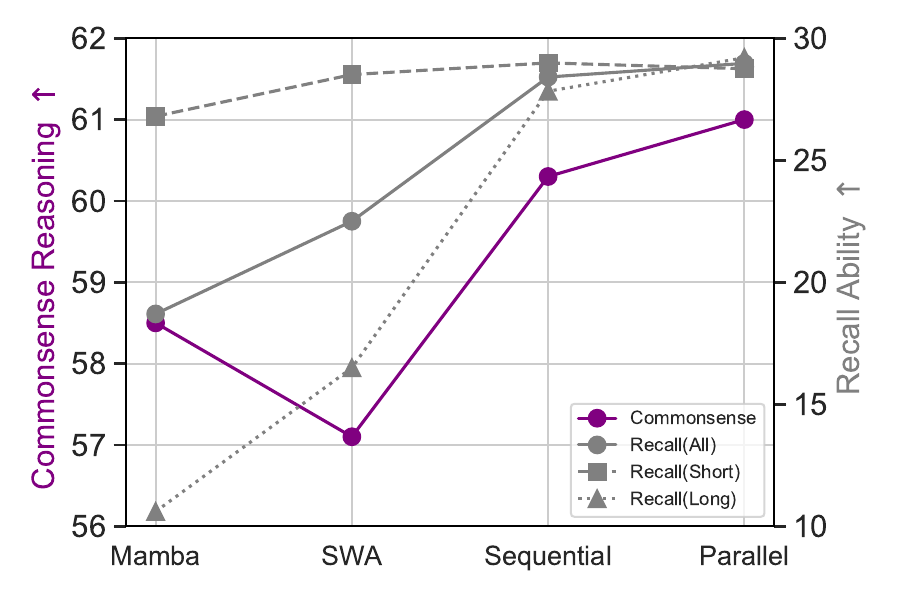}
    \caption{Performance of best performing 1.3B scale models from each architecture in commonsense reasoning and recall ability, where divided by length of context.} 
    \label{app_fig:short_long_comparison_1.3B}
    \end{minipage}
\end{figure}
\subsection{Performance at the 1.3B Scale} \label{app_subsec: 1.3B}

Table~\ref{table: 1.3B} presents the performance of models at the 1.3B parameter scale. Due to computational constraints, we limited our experiments to configurations that demonstrated strong performance at the 430M scale.
We observe consistent trends across both scales. Hybrid models outperform their non-hybrid counterparts. Among sequential architectures, the MFSF model achieves the best performance. Additionally, parallel architectures that use merge-attention layers for fusion generally yield the highest performance.
We also observe a similar pattern from 430M scale~(Figure~\ref{fig:short_long_comparison}) when comparing short- vs. long-sequence settings in recall ability in 1.3B scale~(Figure~\ref{app_fig:short_long_comparison_1.3B}).

\begin{table*}[h]
\centering
\fontsize{6.5}{10} \selectfont
    \begin{tabular}{c|ccccc|c|cccccccc|c}
    \toprule
    &  \multicolumn{6}{c}{Commonsense Reasoning} |& \multicolumn{9}{c}{Recall Ability} \\ 
    \midrule
    Model Type& LAM. & Hella. & PIQA & ARC & Wino. & \textit{\textbf{Avg.}} & NQ-S & NQ-M & NQ-L & Drop & FDA & SWDE & TQA & SQD  &\textit{\textbf{Avg.}}\\
    \midrule
    M & 45.3	&52.7	&72.1	&	67.6	&54.9	&58.5 & 15.8&	13.0&	10.6&	16.9	&4.1	&14.8&	50.8&	23.6 & 18.7\\
    \midrule
    SF & 47.2	&49.8&	69.5	&	65.9	&53.4	&57.1 & 18.0&	16.0	&10.7&	19.1	&10.3	&29.0	&52.0	&24.9&22.5\\
    \midrule
    \multicolumn{6}{l}{\textbf{Sequential Hybrid}} \\
    \midrule
    MS &48.9	&48.8	&69.9	&	65.3	&54.9&57.6 & 17.9&	15.4	&10.3&	19.3	&45.9&	26.4	&53.8&	23.1&26.5\\
    MFSF & 52.9	&52.6&	71.9&	68.5	&55.4	&60.3 & 17.6&	15.6&	10.9&	20.0	&46.2	&38.6	&53.7&	24.7& 28.4\\
    \midrule
    \multicolumn{6}{l}{\textbf{Parallel Hybrid}} \\
    \midrule   
    MF|SF (Avg)&53.5&	51.9	&71.4&	64.1	&56.1&	59.4 &19.1	&16.0	&12.4	&18.6&	47.8	&35.8&	53.3	&24.6 & 28.5\\ 
    MF|SF (MergeAttn) &54.4&	53.7&71.7	&	68.0	&57.4& \textbf{61.0} & 17.9&16.9	&11.8&	19.4	&48.4&	39.7&	51.7&	26.0 & \textbf{29.0}\\ 
    \bottomrule
    \end{tabular}
\caption
     {
   Model performance at the 1.3B scale. Due to computational constraints, we evaluate only those configurations that performed well at the 430M scale.
    Model Type: $M$ = Mamba, $S$ = SWA, $F$ = FF layer. The order reflects the design sequence within each block. In parallel hybrids, "|" denotes parallel branches (e.g., M|SF means Mamba on one side, SWA+FF on the other).
    Tasks: LAM. = LAMBADA-OpenAI, Hella. = HellaSwag, ARC = ARC-Easy, Wino. = Winogrande, NQ-S = NQ-512, NQ-M = NQ-1024, NQ-L = NQ-2048, TQA = TriviaQA, SQD = SQuAD.
    Bold indicates the highest average performance. In both cases, the best models use hybrid architectures with merge-attention.
     } 
\label{table: 1.3B}
\end{table*}

\subsection{Similarity between SWA and Mamba Output Embeddings in Hybrid Models}
\label{app_subsec: similarity btw output}
We observe that sequential hybrids exhibit high similarity between SWA and Mamba outputs, especially in the larger 1.3B model, while parallel hybrids show much lower similarity, particularly in early and middle layers.
This difference arises from the design: sequential hybrids pass outputs from one component to the next, naturally aligning their embedding distributions. In contrast, parallel hybrids process the same input independently, with their outputs aggregated later, leading to more distinct representations. 

This structural difference impacts performance. 
Sequential hybrids maintain a consistent representational space, enabling stable training and strong results on tasks requiring commonsense reasoning or handling shorter contexts. 
However, they struggle with longer-context tasks that require richer representations. 
Parallel hybrids, while more sensitive to aggregation strategies due to the divergence in output spaces, can achieve better performance on complex tasks when trained effectively by leveraging the complementary strengths of both components.

\subsection{Identifying Critical Components in Hybrid Blocks} 
\label{app_subsec: critical_layer}

Figure~\ref{fig:removal_layer} presents the performance reduction rates (y-axis) for commonsense tasks (left) and recall tasks (right) as a function of the block removed (x-axis, represented as the percentage depth from the final block). Across most configurations, the removal of the first block results in the highest performance degradation, indicating that early blocks are typically the most critical. This trend is particularly pronounced in recall tasks, where removing the first block often leads to performance drops of around 90\%.

To better understand the importance of components within each block, we analyze how the removal of specific subcomponents affects performance across architectures. In sequential architectures, the first subcomponent in each block plays the most important role. In contrast, in parallel architectures, the aggregation mechanism, rather than individual components like Mamba or SWA, is the most impactful.
In more details, in sequential architectures, such as MFSF, where the Mamba layer is placed first, removing this initial layer leads to significant degradation, while removing the SWA layer has a milder effect. Conversely, in SFMF, which places the SWA layer first, the most substantial drop occurs when the SWA layer is removed, with the Mamba layer being less impactful (Figure~\ref{app_fig:removal_layer}). 
These results suggest that the position of the layer (i.e., being the first) has a greater influence on performance than the specific type of layer (Mamba vs. SWA).
For parallel architectures, the impact of removing individual Mamba or SWA layers is less severe. Instead, the greatest degradation occurs when aggregation mechanisms such as merge-attention or projection layers are changed to a simple average. 

The findings can also be related to the distribution shift caused by each component.
Sequential architectures exhibit the strongest distributional shift in the first component, making it consistently important regardless of whether it is the Mamba or the SWA component. After this initial transformation, subsequent components tend to collapse into similar distributions, reducing their relative impact.
In contrast, in parallel architectures, both the Mamba and SWA components process the same input independently. As a result, the distributional shifts introduced by each path are less pronounced, and the model can still form a reasonable representation of the input even if one component is removed.
However, the aggregation mechanism causes the largest distributional shift in parallel architectures. 
Replacing it with simpler methods, such as averaging, can distort the combined representation from the two components, resulting in significant performance degradation.

\subsection{Calculating average attending weights} \label{app_subsec: attn_weight}
We calculate Mamba's hidden attention maps following \citet{benkish2024decimambaexploringlengthextrapolation}.  
The average attending weight is calculated with a randomly selected 100 samples of the validation set of Slimpajama of a 4k chunk. 
We average over all tokens and all layers.

\section{Dataset Strategies to Enhance Recall}
\label{app:dataset_strategy}

\subsection{Filtering the Paraphrased Dataset} \label{app_subsec: paraphrase_filtering}

We apply a filtering process to the paraphrased dataset based on the following criteria: (1) the model fails to generate a valid question and answer pair, (2) the generated answer is not present in the corresponding paragraph, or (3) the model fails to convert the example into a cloze-style task, such as when the answer does not appear at the end of the sentence. Instances that do not meet these criteria are discarded, and the processing continues with the remaining examples.
For all experiments, we maintained approximately 3k training instances in the training dataset to ensure a fair comparison.

\subsection{Introducing Paraphrased Data: Early vs. Late} \label{app_subsec: paraphrase_order}

We investigate the impact of introducing paraphrased datasets at different stages of pretraining. When added early, performance deteriorates: although training loss decreases steadily, validation loss increases, suggesting overfitting. We hypothesize this is due to the model’s limited language modeling ability in the early stages, making it more sensitive to data quality. Additionally, deduplication plays a critical role in preventing overfitting. In contrast, introducing paraphrased data later in the continual training stage, as the model is stable, we observe that it consistently improves the recall performance.

\begin{table}[t!]
\centering
\fontsize{6.5}{10} \selectfont
    \begin{tabular}{c|cc}
    \toprule
    Training Dataset& Commonsense &  Recall \\
    \midrule
    Original& 51.5&19.1\\
    \midrule
    Based (SQuAD) & 45.3 &22.3\\
    NIAH & 50.4 &21.6\\
    UltraChat& 47.0 &20.6\\
    \textbf{Ours} & 51.2 &20.4\\
   
    \bottomrule
    \end{tabular}
\caption
     {
        Performance on commonsense reasoning and recall ability after training on the datasets listed in the ``Training Dataset'' column.
    } 
\label{app_table: Seq.Hybrid.Balance}
\end{table}

\subsection{Our Dataset Achieves the Best Balance Across Various Training Datasets} \label{app_subsec: balance_hybrid}

Table~\ref{app_table: Seq.Hybrid.Balance} shows the performance on commonsense reasoning and recall ability after training on datasets on the SQuAD dataset from Based, NIAH, UltraChat, and Ours (paraphrased slimpajama dataset). 
Please note that we remove the SQuAD dataset when averaging the recall ability.

\begin{figure}[t!]
    \begin{minipage}[b]{1.0\linewidth}
    \centering
    \includegraphics[width=1.0\linewidth]{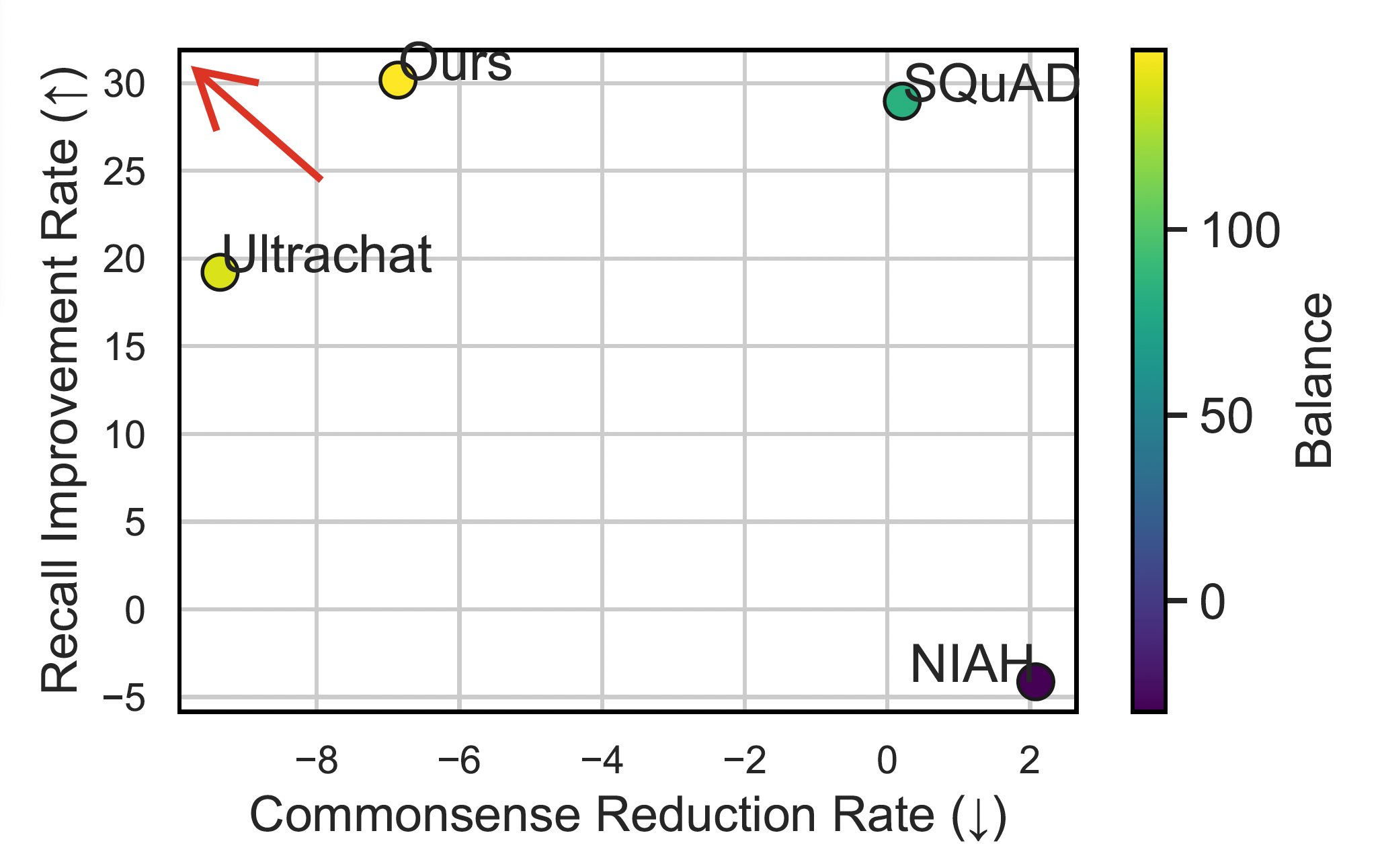}
    \end{minipage}
\caption{
The upper-left region (indicated by the \textcolor{red}{red} arrow) represents the optimal balance between recall improvement and commonsense degradation. When training a pretrained 430M Mamba across various datasets, models trained on our dataset (paraphrased SlimPajama) consistently achieve the best balance compared to when training on other datasets.
}
\label{app_fig:mamba_tradeoff}
\end{figure}

\subsection{Ours Also Shows Good Balance on Mamba-Only Models} \label{app_subsec: balance_mamba}

Figure~\ref{app_fig:mamba_tradeoff} illustrates that models trained on our dataset (paraphrased SlimPajama) tend to achieve an optimal balance, compared to those trained on alternative datasets such as NIAH, Based, or UltraChat, when finetuned on top of the pretrained 430M Mamba model.

Along with hybrid models, we observe notable improvements in recall ability with minimal or no degradation in commonsense reasoning or language modeling performance when training non-hybrid models (models using only Mamba or only SWA layers) with a dataset of 4k sequence length and 40 million tokens.
The mamba-only model showed a recall improvement rate of 29.5\%, whereas the SWA-only model showed a more modest improvement of 17.7\%. This suggests that the Mamba-only model, despite initially exhibiting weak recall performance due to underdeveloped recall capabilities during pretraining, has significant potential for recall when further trained. 
Prior to our additional training, the SWA-only model outperformed the Mamba-only model in recall (SWA: 13.8, Mamba: 12.5). However, after training, the Mamba-only model learned to better retain and recall information, resulting in a recall performance of 17.7, surpassing that of the SWA-only model (16.8).
Furthermore, this improvement in recall did not come at the cost of commonsense reasoning. The Mamba-only model shows a 6.87\% increase in commonsense reasoning, whereas the SWA-only model shows a 2.96\% decline in commonsense reasoning ability. 
These results suggest that our method not only benefits hybrid models but also improves the performance of various model architectures, particularly those utilizing Mamba layers.

\begin{table*}[h]
\centering
\fontsize{6.5}{10} \selectfont
    \begin{tabular}{c|c|ccccc|c|ccccccc|c}
    \toprule
    &  \multicolumn{6}{c}{Commonsense Reasoning} |& \multicolumn{9}{c}{Recall Ability} \\ 
    \midrule
    Base Model & Type & LAM. & Hella. & PIQA & ARC & Wino. & \textit{\textbf{Avg.}} & NQ-S & NQ-M & NQ-L & Drop & FDA & SWDE & TQA  &\textit{\textbf{Avg.}}\\
    \midrule
    \multirow{2}{*}{Mamba} 
    &  & 69.1 & 49.5 & 75.3 & 64.1 & 63.2 & 63.7 & 31.0 & 28.1 & 21.7 & 20.9 & 29.6 & 41.0 & 64.6 & 33.8\\ 
    & + Ours & 67.0 & 64.8 &76.0 & 68.3 & 62.4 & 65.4& 41.0 & 37.3 & 27.5 & 31.5& 32.2 &41.0 & 71.4 & 40.3\\
    \midrule
    \multirow{2}{*}{Mamba-U} 
    & & 67.0 & 70.5 & 78.6 & 65.9 & 65.2 & 65.6 & 36.3	&35.0&	27.7	&25.7	&34.3	&50.1	&70.5& 39.9\\ 
    & + Ours & 65.9&	69.7	&78.3	&69.8	&64.0 &	66.9 &42.6	&39.4	&30.8	&30.8	&33.6	&52.4&	74.8& 43.5\\ 
    \midrule
    \multirow{2}{*}{Mamba-Z}
    & &67.9	&71.2	&78.4	&66.2	&65.0 &	65.6 & 36.8	&35.1&	27.8	&26.1	&32.8&	51.6&	70.4&40.1\\ 
    & + Ours & 66.8&	69.7&	77.8&	70.2	&67.8&	69.0 &42.4	&38.5&	30.6	&31.3&	34.2	&34.9	&74.3&40.9 \\ 
    \bottomrule
    \end{tabular}
\caption
     {
Performance of Mamba-2.8B when continually trained on our paraphrased dataset, evaluated across different base model variants. We observe consistent improvements in both commonsense reasoning and recall capabilities, with gains more pronounced for stronger base models (e.g., Mamba). "Mamba-U" and "Mamba-Z" refer to Mamba-2.8B-UltraChat and Mamba-2.8B-Zephyr, respectively.
     } 
\label{app_table: paraphrase_basemodel}
\end{table*}

\subsection{Comparison with DeciMamba} \label{app_subsec: decimamba}
We evaluate performance on the LongBench dataset to compare our training approach with DeciMamba, using the same base model (Table~\ref{app_table: decimamba}). Model trained with our dataset consistently yields stronger results, particularly on QA datasets, with an average improvement of +8.1 points.

\subsection{Generalization Across Different Mamba-2.8B Variants} \label{app_subsec: basemodel}
Table~\ref{app_table: paraphrase_basemodel} presents the performance of various base models trained using our paraphrased dataset.
To ensure a fair comparison, we evaluate three variants of the Mamba-2.8B model: \href{https://huggingface.co/state-spaces/mamba-2.8b}{Mamba-2.8B}, \href{https://huggingface.co/xiuyul/mamba-2.8b-ultrachat}{Mamba-2.8B-Ultrachat}, and \href{https://huggingface.co/xiuyul/mamba-2.8b-zephyr}{Mamba-2.8B-Zephyr}.
Our results show consistent improvements in both commonsense reasoning and recall performance when using the paraphrased dataset. Notably, the gains are most pronounced when using the original Mamba-2.8B model as the base, suggesting that models with fewer prior instruction-tuning steps may benefit more from paraphrased augmentation.

\subsection{Length of Training Dataset} \label{app_subsec: paraphrase_training_dataset_length}
\begin{figure*}[t!]
    \begin{minipage}[b]{1.0\textwidth}
    \centering
    \includegraphics[width=0.9\textwidth]{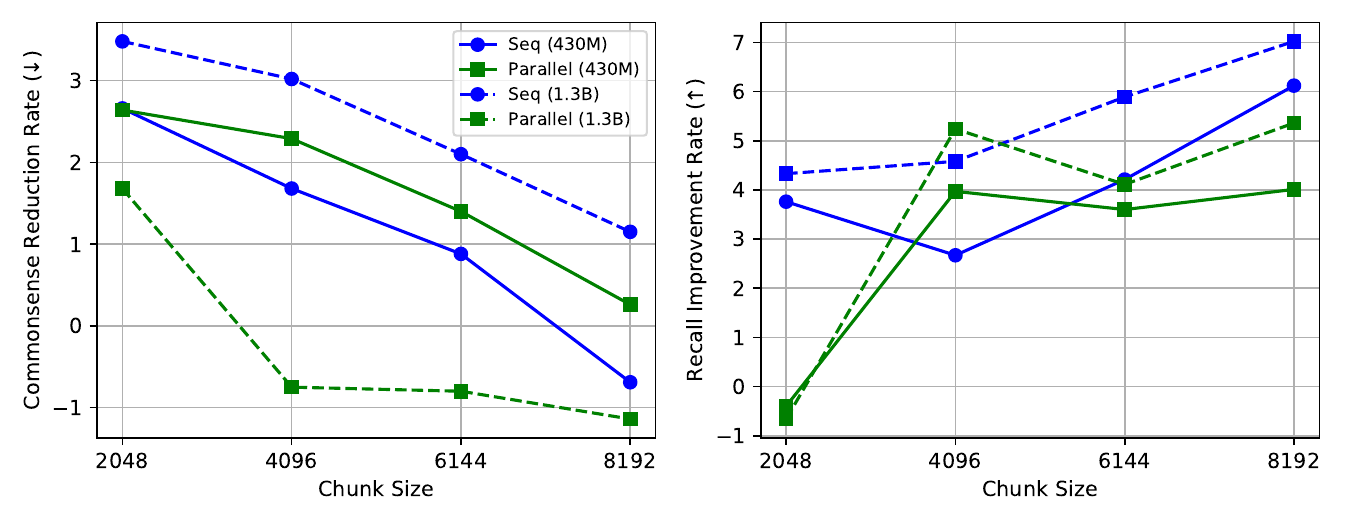}
    \end{minipage}
\caption{Commonsense reasoning reduction rate(left) and recall improvement rate(right) by changing the chunk size of the training dataset (x-axis).} 
\label{app_fig:dataloader_length}
\end{figure*}

\begin{figure}[t!]
    \begin{minipage}[b]{1.0\linewidth}
    \centering
    \includegraphics[width=0.9\linewidth]{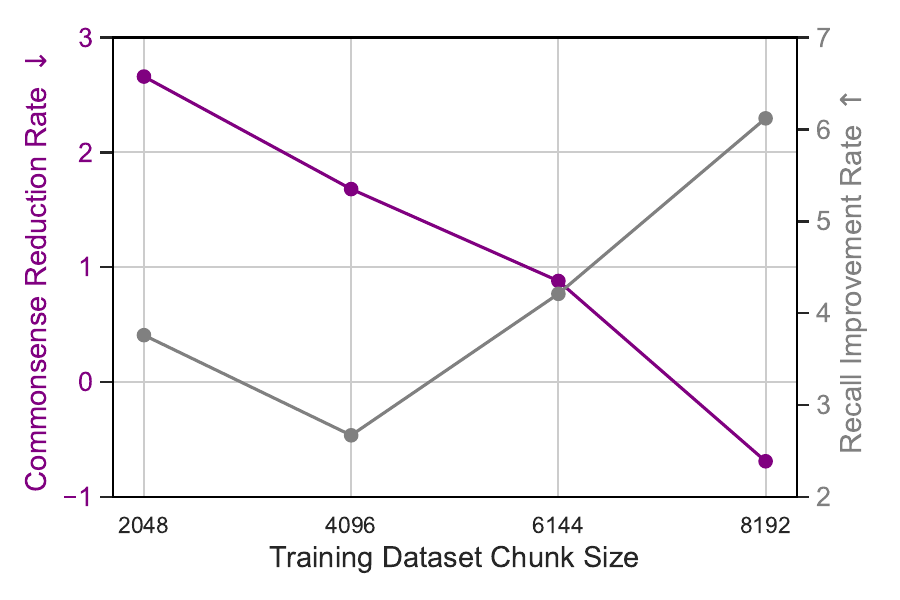}
    \end{minipage}
\caption{\textcolor{purple}{Commonsense reasoning reduction rate} and \textcolor{gray}{recall improvement rate} by changing the chunk size of the training dataset (x-axis).} 
\label{fig:dataloader_length}
\end{figure}

\begin{table}[t!]
\centering
\fontsize{6.5}{10} \selectfont
    \begin{tabular}{c|c|ccc}
    \toprule
    Length & Commonsense & Recall (Short) & Recall (Long) & Recall (All) \\
    \midrule
    Original & 51.5 & 22.33	& 16.35	& 19.34\\
    \midrule
    2k &50.1 & 24.3&	16.4&	19.8\\
    4k & 50.6 & 23.6	&16.5	&19.6\\
    6k & 51.0& 23.7&	17.0	&19.9\\
    8k & 51.9&24.4	&17.3	&20.3\\
    \bottomrule
    \end{tabular}
\caption
     {
        Average commonsense and recall performance for short and long contexts as training dataset length (Length Column) varies in sequential hybrid (MFSF) training.
    } 
\label{app_table: Seq.Hybrid.Balance}
\end{table}

As shown in Figure~\ref{app_fig:dataloader_length}, for both sequential and parallel architectures and various model sizes, \textbf{continually training with longer chunk size result in lower reduction rate on commonsense tasks and higher improvements on recall tasks}.

Upon closer inspection (Table~\ref{app_table: Seq.Hybrid.Balance}), shorter chunk sizes (e.g., 2k) significantly boost performance on short-context recall tasks but lead to notable degradation on long-context tasks. This effect is particularly pronounced in parallel models. We hypothesize that this is because, as shown in Section~\ref{app_subsec: parallel_agg}, parallel hybrid retains layer-wise characteristics more strongly than sequential models.
Additionally, the gap in performance is more substantial for recall tasks (range of around -1\% to 7\%) than for commonsense tasks (range of around -1\% to 3\%).

\begin{table*}[t!]
\centering
\fontsize{6.5}{10} \selectfont
    \begin{tabular}{@{}lcccc@{}}
    \toprule
\textbf{Benchmark} & \textbf{Avg Len} & Mamba & DeciMamba & Mamba + Ours \\
\midrule
2wikimqa & 4887 &  3.9 & \textbf{9.1} & 8.0\\
Hotpotqa & 9151 & 1.5 & 4.5 & \textbf{12.5}\\
Musique & 11214 & 0.9 &  1.7& \textbf{2.3} \\
Narrative QA & 18409 & 0.9 & 1.7 & \textbf{3.5}\\
Qasper & 3619 &5.97 &  \textbf{8.9} & 8.5 \\
Multifield QA & 4559 &  11.2 & 18.6 & \textbf{19.3} \\
GovReport & 8734 &9.8 &  14.9 & \textbf{15.2}\\
QMSum & 10614 &\textbf{8.2} &  7.1&7.3 \\
MultiNews & 2113 &  23.2 & \textbf{24.6} & 23.7 \\
TriviaQA & 8209 &  3.9 &  12.6 &\textbf{36.0} \\
SAMSum & 6258 &  \textbf{8.6}  & 7.3 & 6.9 \\
TREC & 5177 & 0.5& 0.5 & \textbf{27.0} \\
LCC & 1235 &  8.1 &  8.7  & \textbf{8.9}\\
RepoBench-p & 4206 &  7.2 &  \textbf{11.0} & 10.7 \\
Passage Count & 11141 & 0.0 & \textbf{0.5} & 0.0\\
Passage Ret. en & 9289 &  0.0 & 1.5 & \textbf{1.9}\\
\bottomrule
\end{tabular}
\caption{Performance over LongBench. Results of DeciMamba are from the paper~\citep{benkish2024decimambaexploringlengthextrapolation}. \textbf{Mamba+Ours} is model continual trained with our paraphrased dataset on the same base model (instruction-tuned Mamba-2.8b model). Ours tend to show high performance, especially on QA datasets.}
\label{app_table: decimamba}
\end{table*}

\subsection{Number of training dataset} \label{app_subsec: training_dataset_cnt}

\begin{figure}[t!]
    \begin{minipage}[b]{1.0\linewidth}
    \centering
    \includegraphics[width=1.0\linewidth]{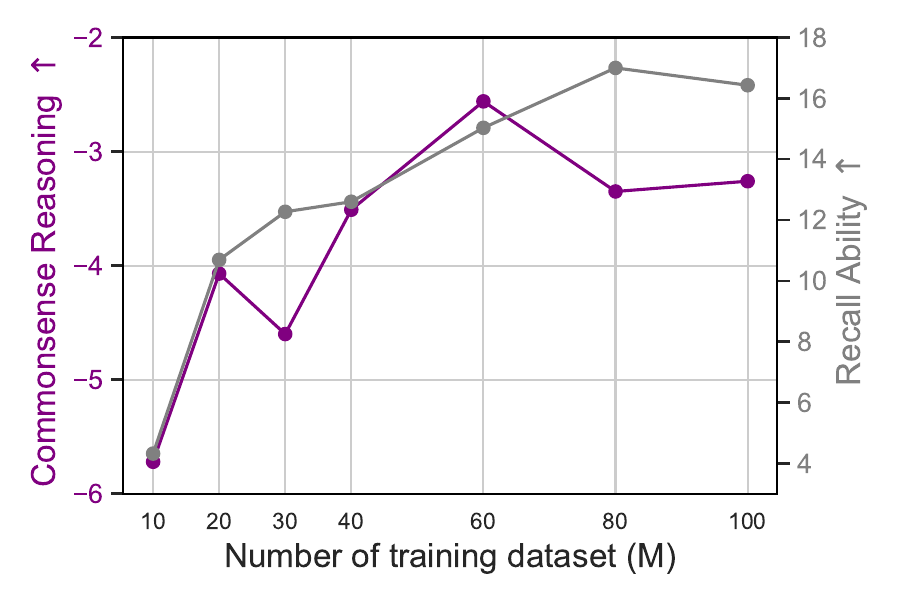}
    \end{minipage}
\caption{\textcolor{purple}{Commonsense reasoning} and \textcolor{gray}{recall ability} when changing the number of training dataset (x-axis).} 
\label{fig:app_dataloader_num}
\end{figure}

Figure~\ref{fig:app_dataloader_num} shows the reduction rate of commonsense performance (left) and the improvement rate of recall performance (right) by the number of training token (x-axis), trained with a chunk size of 4k. 
As the training data size increases, we observe a general improvement in both commonsense and recall performance with convergence of around 80M to 100M tokens.

\end{document}